\documentclass[lettersize,journal]{IEEEtran}
\usepackage{amsmath,amsfonts}
\usepackage{algorithmic}
\usepackage{algorithm}
\usepackage{array}
\usepackage{amsmath}
\usepackage{textcomp}
\usepackage{stfloats}
\usepackage{url}
\usepackage{verbatim}
\usepackage{graphicx}
\usepackage{cite}
\usepackage{epstopdf}
\usepackage{gensymb}
\usepackage{subfigure}
\usepackage{booktabs} 
\usepackage{multirow}
\usepackage{makecell}
\usepackage{tabularx}
\usepackage{xcolor}
\usepackage{bm}
\begin{document}

\title{WLTCL: Wide Field-of-View 3-D LiDAR Truck Compartment Automatic Localization System}

\author{Guodong Sun, Mingjing Li, Dingjie Liu, Mingxuan Liu, Bo Wu and Yang Zhang
\thanks{This work was supported in part by the Hubei province science and technology project of open bidding for selecting the best candidates (2024BEB018), the National Natural Science Foundation of China (Grant 62403186), the National Natural Science Foundation of Hubei Province (2024AFB153). (Corresponding author: Yang Zhang, Bo Wu).}
\thanks{G. Sun, M. Li, D. Liu, M. Liu, and Y. Zhang are with the School of Mechanical Engineering, Hubei University of Technology, Wuhan, 430068, China and the Hubei Key Laboratory of Modern Manufacturing Quality Engineering, Hubei University of Technology, Wuhan, 430068, China (e-mail: sunguodong@hbut.edu.cn; 102210050@hbut.edu.cn; liudingjie@hbut.edu.cn; 102310158@hbut.edu.cn; yzhangcst@hbut.edu.cn).}
\thanks{Bo Wu is with the Shanghai Advanced Research Institute, Chinese Academy of Sciences, Shanghai 201210, China (e-mail:wubo@sari.ac.cn).}
}

\markboth{}
{Shell \MakeLowercase{\textit{et al.}}: A Sample Article Using IEEEtran.cls for IEEE Journals}

\maketitle

\begin{abstract}
As an essential component of logistics automation, the automated loading system is becoming a critical technology for enhancing operational efficiency and safety. Precise automatic positioning of the truck compartment, which serves as the loading area, is the primary step in automated loading. However, existing methods have difficulty adapting to truck compartments of various sizes, do not establish a unified coordinate system for LiDAR and mobile manipulators, and often exhibit reliability issues in cluttered environments. To address these limitations, our study focuses on achieving precise automatic positioning of key points in large, medium, and small fence-style truck compartments in cluttered scenarios. We propose an innovative wide field-of-view 3-D LiDAR vehicle compartment automatic localization system. For vehicles of various sizes, this system leverages the LiDAR to generate high-density point clouds within an extensive field-of-view range. By incorporating parking area constraints, our vehicle point cloud segmentation method more effectively segments vehicle point clouds within the scene. Our compartment key point positioning algorithm utilizes the geometric features of the compartments to accurately locate the corner points, providing stackable spatial regions. Extensive experiments on our collected data and public datasets demonstrate that this system offers reliable positioning accuracy and reduced computational resource consumption, leading to its application and promotion in relevant fields.
\end{abstract}


\section{Introduction}
\IEEEPARstart{I}{ntelligent} logistics systems aim to enhance the automation level of the entire logistics process through the use of smart technologies such as intelligent hardware, robotics, and 3-D vision perception. Loading and unloading, as an essential part of logistics, encompass transportation, storage, and picking operations. It is a key factor in determining the overall efficiency of the logistics process \cite{ref1}. Currently, compared to boxed materials, bagged materials are mostly handled manually, which is considered one of the most labor-intensive operations. Manual handling not only results in low efficiency but also directly increases overall costs. Additionally, some bagged agricultural materials have chemical corrosiveness that can be harmful to human health \cite{ref2}.

Compared with manual palletizing, industrial robots have the advantages of large loading capacity, high speed, and low cost. The material stacking robot stacks the materials at the designated position of the vehicle compartment according to a certain mode, such as in the compartment of a fenced truck, to realize the storage, handling, loading and unloading, transportation, and other logistics activities of bagged materials. The existing palletizing robot loads the goods into the compartment according to the preset loading target point. This requires that before automatic loading, the truck driver can adjust the truck compartment to stop at the specified position in the specified attitude. Once the parking position or compartment attitude of the truck is inconsistent with that of the truck, the automatic loading system cannot complete automatic loading. However, in many cases, loading areas are extremely narrow, making it difficult for drivers to adjust the truck's position and orientation as needed. This limitation hinders current teach-based industrial robot loading systems, which struggle to determine the truck's position and improve loading efficiency. Therefore, the key to enhancing automatic loading efficiency is accurately locating the truck's compartment in various parking orientations.

\begin{figure}[!t]
	\centering 	
        \subfigure[]{ 
		\includegraphics[width=1.6in]{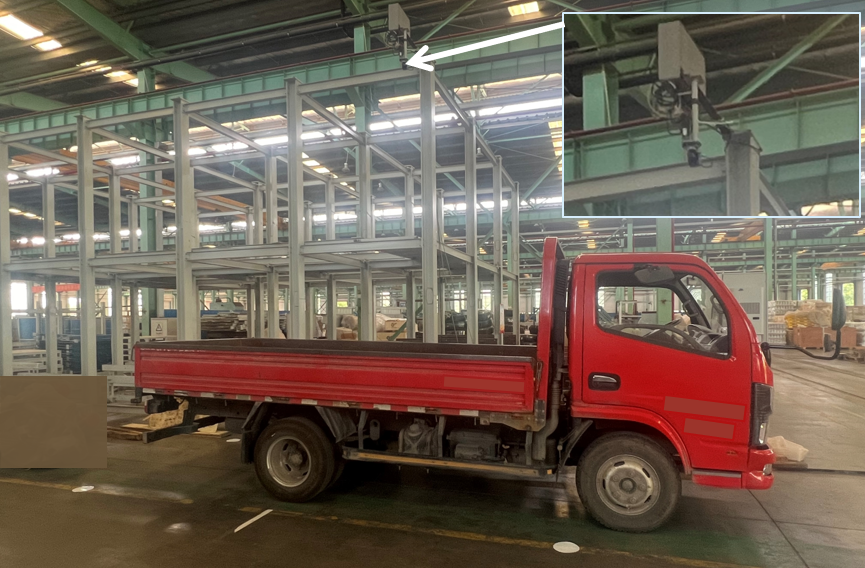}
	}\hspace{-0.7em}
        \subfigure[]{
            \includegraphics[width=1.6in]{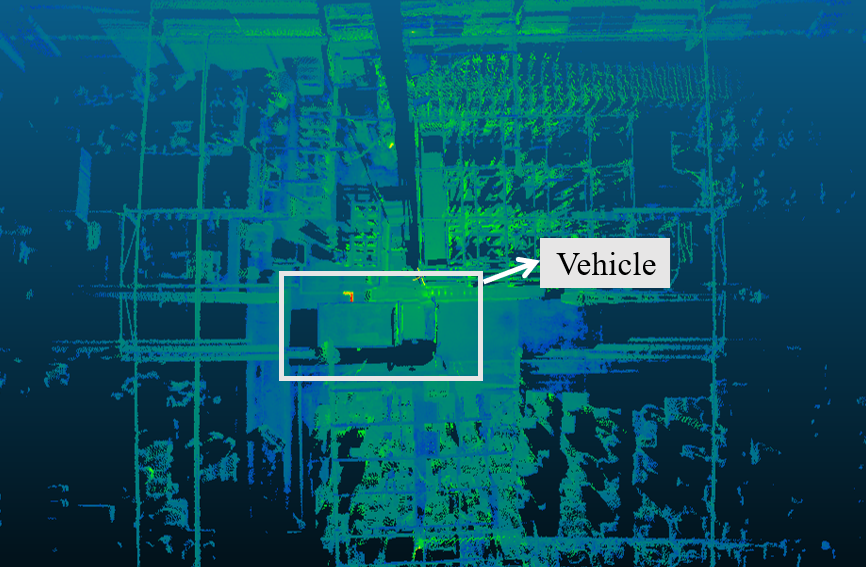}
        }
        \subfigure[]{
            \includegraphics[width=1.63in]{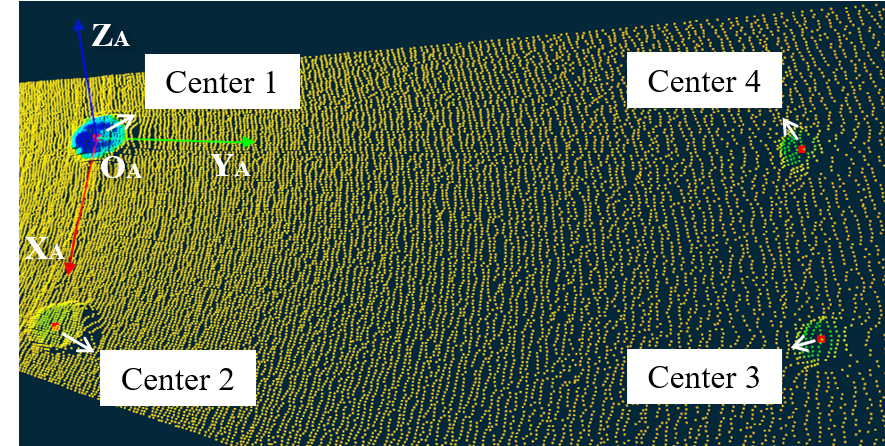}
        }\hspace{-0.7em}
        \subfigure[]{
            \includegraphics[width=1.6in]{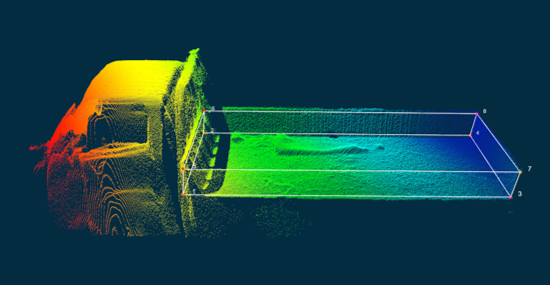}
        }
	\caption{\textcolor{black}{Localization system execution steps. (a) The vehicle enters the parking area. (b) Wide-field-of-view LiDAR scans the scene point cloud. (c) Establish world coordinate system. (d) Vehicle point cloud segmentation and compartment key point localization.}}
	\label{abstract}
\end{figure}
Previous studies have rarely attempted to automatically locate the position of a truck compartment as done in this study. Existing methods can be divided into machine vision and LiDAR measurement. Methods based on machine vision have the advantages of fast measurement speed and high efficiency in practical application. Still, they are easily affected by light, and dust in the loading environment, resulting in reduced measurement reliability \cite{ref3, ref4, ref5}. Additionally, the accuracy of image detection technology is difficult to ensure in large-scale scenarios. Light Detection and Ranging (LiDAR) can provide stable point clouds under varying lighting conditions and is effective in large-scale scenarios. However, ordinary 3-D LiDARs have a small field of view (FOV), making it difficult to obtain a complete point cloud for large vehicles with a single LiDAR \cite{ref6, ref7}. Using multiple LiDARs can solve this problem but significantly increase the equipment cost.

To address the aforementioned challenges, this paper introduces a wide-field-of-view LiDAR-based system comprising hardware and software modules. The hardware component introduces a rotating LiDAR designed to generate high-density vehicle point clouds over a wide field of view, suitable for multi-sized vehicles and environments with uneven lighting. The software component presents a method for establishing a global coordinate system and simultaneously segmenting the vehicle point cloud. Additionally, a vehicle key point localization algorithm is proposed to directly process the segmented vehicle point cloud and output accurate key points of the truck compartment, thereby providing a feasible solution for identifying the loading and unloading space. 

Our main contributions can be summarized as follows:
\begin{enumerate}
    \item We explore a wide field-of-view 3-D LiDAR truck compartment automatic localization system to enhance the automation and reliability of key point localization in fenced truck compartments.
    \item  \textcolor{black}{For cluttered scenes, we design a method to locate the corner points of the parking area, which is used to segment the vehicle point cloud and provide unified 3-D coordinates for the mobile manipulator.}
    \item We propose a key point localization algorithm for truck compartments that leverages vehicle geometric features and prior information, exhibiting robustness to point cloud loss and foreign object interference.
    \item Extensive experiments are conducted on both self-collected and public datasets to validate the stability and reliability of the system.
\end{enumerate}

The remainder of this paper is organized as follows: Section \uppercase\expandafter{\romannumeral2} discusses related work. Section \uppercase\expandafter{\romannumeral3} introduces the hardware principles and software algorithms. Section \uppercase\expandafter{\romannumeral4} presents the experimental results. In the end, Section \uppercase\expandafter{\romannumeral5} summarizes the work of this paper.

\section{Related Works}
\subsection{Truck Compartment Localization}
Despite the relatively limited number of prior studies, they have nonetheless provided valuable insights for this research. Ma et al. \cite{ref3}  used a stereo camera to obtain complete images of trucks and applied the FCN-ResNet50 network with a CBAM attention mechanism to segment the truck images and locate the compartment area and its four corner points. Mrovlje et al. \cite{ref4} utilized a binocular vision system to capture images of trucks, relying on a streamlined image recognition algorithm to accurately locate the trucks in the captured images and thereby determine their spatial position. Zhang et al. \cite{ref5} achieved compartment region recognition and center point localization by extracting edge lines and corner coordinates from depth images. Machine vision-based methods are susceptible to light and dust in the loading environment, resulting in reduced measurement reliability.

Zou et al. \cite{ref6} utilized LiDAR to collect vehicle point cloud data and performed spatial geometric calculations by segmenting and fitting planar layers to accurately determine the position of the compartment. Wei et al. \cite{ref7} employed a 3-D LiDAR to obtain point cloud data of object surfaces and proposed a statistical polyline edge detection algorithm to determine truck dimensions. However, ordinary 3-D LiDARs have a small field of view (FOV), making it difficult to obtain a complete point cloud for large vehicles with a single LiDAR. Using multiple LiDARs can solve this problem but significantly increase the equipment cost. Therefore, we opted to use a rotating platform to drive the 2-D LiDAR, which provides a wider field of view while reducing equipment costs.

\subsection{3-D Line Segment Detection}
The 3-D line features are among the most common and significant characteristics of fence trucks. Accurately and efficiently extracting these 3-D line features is crucial for precise compartment key point localization. Poreba et al. \cite{ref8} employed the random sampling consistency algorithm \cite{ref9,ref10} and region-growing algorithms to extract planes from point clouds, deriving 3-D line segments from their intersections \cite{ref11,ref12}. Moghadam et al. \cite{ref13} employed the local principal component analysis (PCA) algorithm \cite{ref14} to extract edge points with depth discontinuities, followed by clustering and fitting of the lines. Luo et al. \cite{ref15} incorporated a line position refinement process that accounts for the uncertainties of spatial lines to construct more accurate maps. Zhao et al. \cite{ref16} utilized shared EdgeConv encoder layers to jointly train two heads for segmentation and descriptor to achieve line segmentation and description of LiDAR point clouds. 

Kong et al.\cite{ref17} used LiDAR point clouds for clustering and fitting the intersection lines of planes to achieve precise measurements of building rooftops. Lin et al.\cite{ref18} employed the dual consistency constraint of edge point coordinates and horizontal lines to fit line segments, enabling efficient line segment detection. Du et al.\cite{ref19} introduced a novel line segment grouping method, completing rapid multi-plane segmentation of LiDAR point clouds by alternately searching for candidate plane seeds and adjacent line segments.

Lu et al. \cite{ref20} introduced a simple yet effective large-scale 3-D line detection algorithm for unorganized point clouds. This method was based on point cloud segmentation and 2-D line detection, followed by post-processing to remove outliers. It is highly effective for structured scenes. Our localization method is inspired by the concept of point cloud segmentation and 2-D line detection as described in \cite{ref20}. Building on this foundation, we designed a compartment key point localization algorithm that achieves efficient and reliable estimation of the compartment position.

\section{Proposed System}
\begin{figure*}[!t]
    \centering
    \includegraphics[width=6.5in]{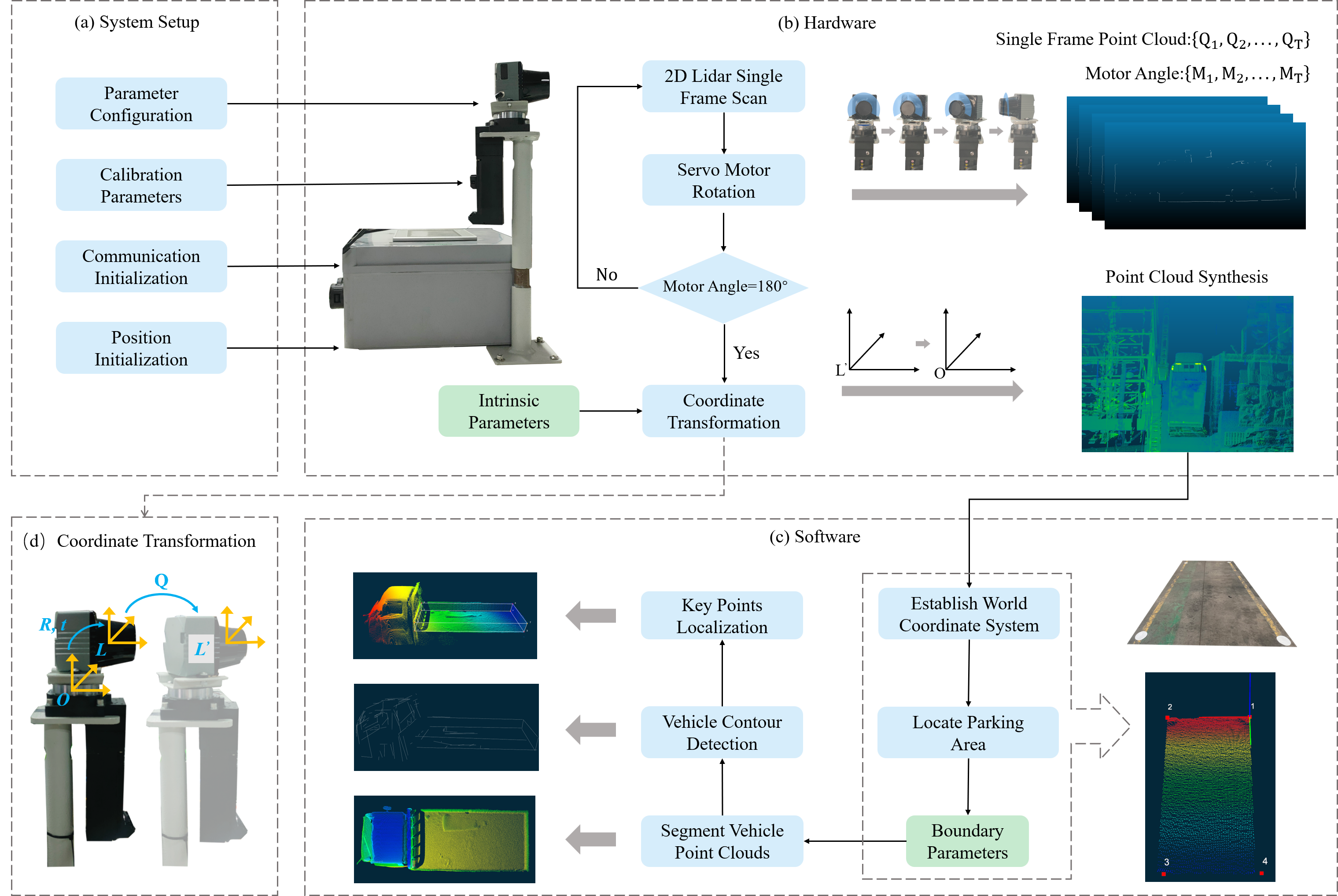}
    \caption{System overview. First, during the startup phase, the system needs to load the parameter configuration of the software and hardware modules. Second, the vehicle enters the parking area, and the wide-field-of-view LiDAR starts scanning to generate the scene point cloud. Then, the system establishes a loading coordinate system and segments the vehicle point cloud. Finally, the compartment key point localization algorithm identifies the key points of the compartment and outputs the loadable area, completing the compartment localization.}
    \label{pipline}
\end{figure*}

The overall framework, depicted in Fig. \ref{pipline}, mainly comprises the LiDAR hardware measurement module and the software module for compartment key point localization. Initially, during the system startup phase, parameter configuration for both hardware and software modules is conducted. Subsequently, the LiDAR is activated to scan and generate a point cloud of the scene. Following this, the system establishes the loading coordinate system, locates the parking area, and segments the point cloud of the vehicle according to the boundary parameters. Finally, the positioning algorithm locates the compartment key points and calculates the space where the goods can be stacked.

\begin{table}[!t]
	\centering
	\renewcommand{\arraystretch}{1.17}
	\caption{Notations of System}
	\footnotesize
	\begin{tabular}{lp{6.5cm}}
		\toprule
        \toprule
		\textbf{Symbol}   &\textbf{\makecell[c]{Explanation}} \\
		\midrule
		$L$  & {Lidar coordinate system.} \\ 
        $L^{\prime}$ & {Coordinate system after rotation.} \\
        $\theta$ & {Azimuth angle of the ith ray of the LiDAR.} \\
        $O$  & {Coordinate frame of the rotating platform (origin).} \\
        $^{O}P$ & {Point with respect to $O$.} \\
        $q$ & {Point with respect to $L$.} \\
        $\phi$ & {Rotation angle measurement from the rotating platform.} \\
        $W$ & {Euler angles between $O$ and $L$: Yaw, Pitch, Roll.} \\
        $Q$ & {Rotation matrix indicating rotation by the rotating platform.} \\
        $R_{w}$ & {Rotation matrix between $O$ and $L$ indicating rotation by $w$.} \\
        $T_t$ & {Translation matrix between $O$ and $L$ indicating translation by $t$.} \\
        $^{O}_{L}T$ & {Homogeneous transformation matrix of L with respect to O.} \\
        $A$ & {World coordinate system.} \\
        $^{A}P$ & {Point with respect to $A$.} \\
        $P_{AORG}$ & {World coordinate system origin.} \\
        $P_{OORG}$ & {Origin of the $O$.} \\
        $\overline{OX_{A}}$ & {X-axis of coordinate system A.} \\
        $\overline{OY_{A}}$ & {Y-axis of coordinate system A.} \\
        $\overline{OZ_{A}}$ & {Z-axis of coordinate system A.} \\
        $^{O}_{A}T$ & {Homogeneous transformation matrix of A relative to O.} \\
        $^{A}_{O}T$ & {Homogeneous transformation matrix of O relative to A.} \\
        $k$ & {The number of neighbors per data point in the segmentation plane.} \\
        $\textcolor{black}{\delta}$ & {\textcolor{black}{Constant angle threshold for region growth in the segmentation plane.}} \\
        $C_i$ & {Cosine similarity between line segments.} \\
        $D_i$ & {Distance between line segments.} \\
		\bottomrule
	\end{tabular}
	\label{NOMENCLATURE}
\end{table}

\IEEEpubidadjcol
\subsection{Ranging and Scanning Principles}
The task of the measurement system is to obtain the three-dimensional coordinates $(^{O}P)$ of the measured object relative to the reference coordinate system. We fix the coordinate system $(O)$ of the measurement system to the rotation center of the rotating platform, and then the problem of obtaining the three-dimensional coordinate $(^{O}P)$ is defined as solving the transformation relationship between the rotating LiDAR coordinate system $\left( L^{\prime} \right)$ and the reference coordinate system $(O)$, as shown in Fig. \ref{pipline}\text{(d)}. The functional relationships available in the measurement system are as follows:
\subsubsection{2-D LiDAR measurement}
\textcolor{black}{The 2-D LiDAR system emits multiple pulsed lasers using a built-in rotating mirror. It measures the target distance $r$ and the corresponding azimuth angle \text{$\theta$}. The coordinate $(q)$ of the measured object, relative to the LiDAR coordinate system $(L)$, is given by $q=r{\left [\sin \theta  \enspace \cos \theta \enspace 0 \right ]}^{T}$.}
\subsubsection{Rotating platform measurement}
The rotating platform is equipped with an angle encoder that drives the 2-D LiDAR to rotate around a specified axis to achieve 3-D scanning and measurement. The value of the angle encoder $(\phi)$ is used to calculate the rotation matrix $(Q)$, which describes the transformation from the rotated LiDAR coordinate system $\left( L^{\prime} \right)$ to the LiDAR coordinate system $(L)$.

According to the functional relationships mentioned above, the 3-D coordinates of point $(^{O}P)$ can be obtained by converting 2-D point $(q)$ into a point in the coordinate system $(L)$, and then converting the point coordinates from $(L)$ to $\left( L^{\prime} \right)$. The above transformation constitutes the measurement equation
$^{O}P=Q({R}_{w}\cdot q+t)$,
where the transformation matrix $(Q)$ is represented as follows:
\begin{align}\label{eq:model}
    Q=\begin{bmatrix}
    \cos(\phi) & 0 & \sin(\phi) \\ 
    0 & 1 & 0 \\ 
    -\sin(\phi) & 0 & \cos(\phi)
    \end{bmatrix}, \notag \\ 
    {R}_{w}={R}_{z}({w}_{z}){R}_{y}({w}_{y}){R}_{x}({w}_{x}), \notag \\
    T_t=[T_{x} \enspace T_{y}  \enspace T_{z} ],
\end{align}
where $w=[{w}_{x}\enspace{w}_{y}\enspace{w}_{z}]=[roll\enspace \textcolor{black}{pitch} \enspace yaw]$,
\begin{align}
    {R}_{x}({w}_{x})=\begin{bmatrix}
1 &0 &0 \\ 
0 &\cos({w}_{x}) &-\sin({w}_{x}) \\ 
0 &\sin({w}_{x}) &\cos({w}_{x})
\end{bmatrix}, \notag \\ 
    {R}_{y}({w}_{y})=\begin{bmatrix}
\cos({w}_{y}) &0  &\sin({w}_{y})  \\ 
0 &1   & 0 \\ 
-\sin({w}_{y}) &0  & \cos({w}_{y})
\end{bmatrix}, \notag \\ 
    {R}_{z}({w}_{z})=\begin{bmatrix}
\cos({w}_{z}) &-\sin({w}_{z})  &0  \\ 
\sin({w}_{z}) &\cos({w}_{z})  & 0 \\ 
0 &0  & 1
\end{bmatrix}.
\end{align}

\textcolor{black}{The method in [21] can estimate accurate calibration parameters without requiring additional hardware. We adopted this approach, utilizing standard planar calibration parameters $R_{w}$ and $T_t$.} Initially, the rotation platform is rotated 360°, and the 2-D LiDAR scans the calibration plane from different positions and orientations on its left and right sides. The range data from these calibration planes are then segmented, resulting in multiple datasets $\{r_{i}, \theta_{i}, \phi_{i}\}$. By utilizing the properties that the normal vector of the plane is perpendicular to the lines within the plane and the coplanarity of the plane, we determine the calibration parameters. The rotational cost function and translational cost function of the plane are defined as follows:
\begin{align}
&\boldsymbol{E}_{rot}(R_{w}) = \left[\begin{array}{llll}
\hat{\boldsymbol{d}}_1 & \hat{\boldsymbol{d}}_2 & \ldots & \hat{\boldsymbol{d}}_N
\end{array}\right]^T \hat{\boldsymbol{n}}, \notag \\ 
&\boldsymbol{E}_{trans}(t) = \left[\begin{array}{c}
\hat{\boldsymbol{p}}_1^T \hat{\boldsymbol{n}} \\
\hat{\boldsymbol{p}}_2^T \hat{\boldsymbol{n}} \\
\vdots \\
\hat{\boldsymbol{p}}_M^T \hat{\boldsymbol{n}}
\end{array}\right] - \left[\begin{array}{c}
\mu \\
\mu \\
\vdots \\
\mu
\end{array}\right], \notag \\ 
&\mu = \frac{1}{M} \sum_{i=1}^M \hat{\boldsymbol{p}}_i^T \hat{\boldsymbol{n}},
\end{align}
where $P_{i}$ represents points within the plane, $n$ is the normal vector of plane, and $d_{i}$ denotes the direction of each ray forming the plane. The least-squares solution of the cost function can be obtained using iterative methods such as the Levenberg-Marquardt (LM) algorithm\cite{ref22}.

\begin{figure}[!t]
	\centering 	
        \subfigure[]{
		\includegraphics[width=3.2in]{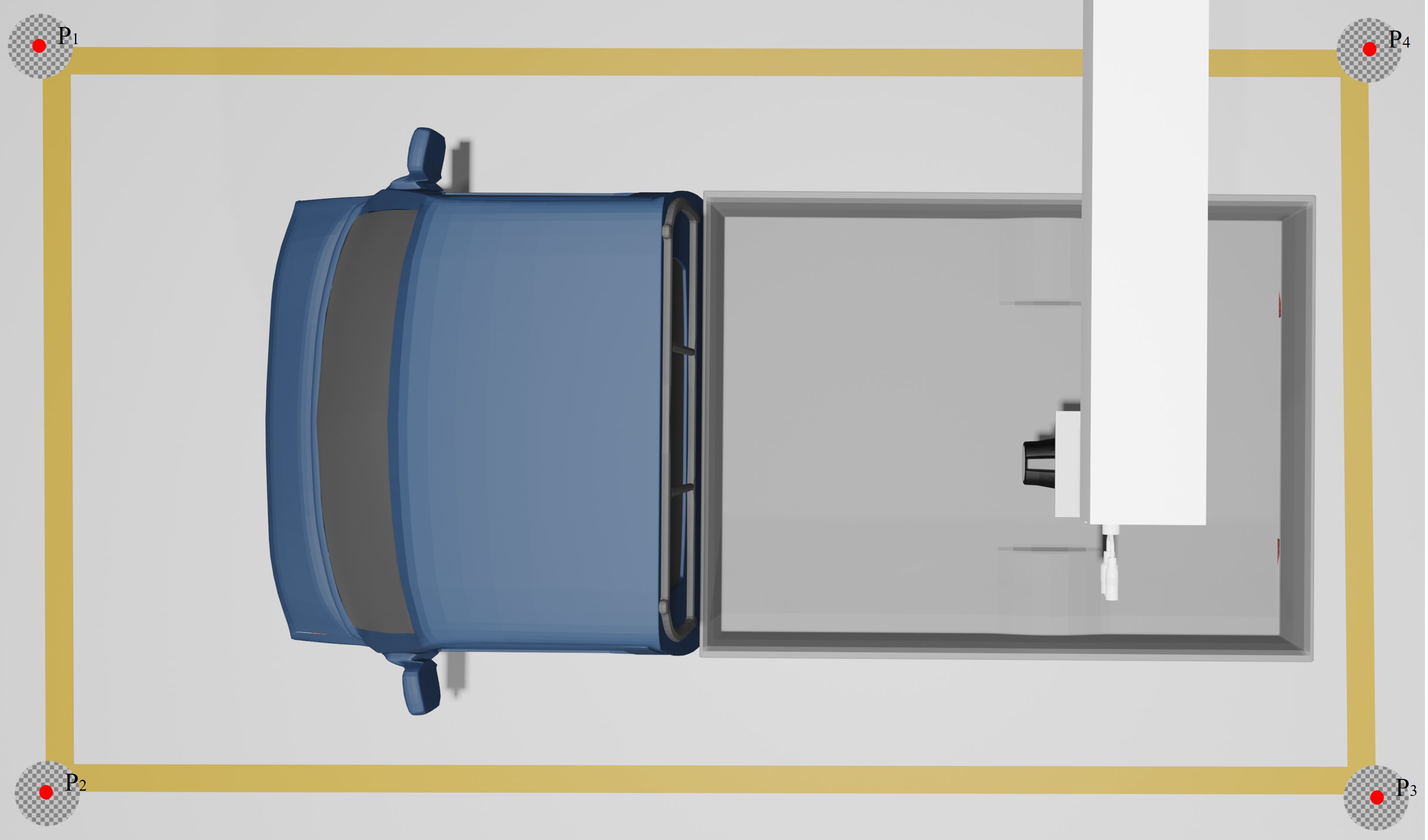}
	}\hspace{-0.7em}
        \subfigure[]{
            \includegraphics[width=3.2in]{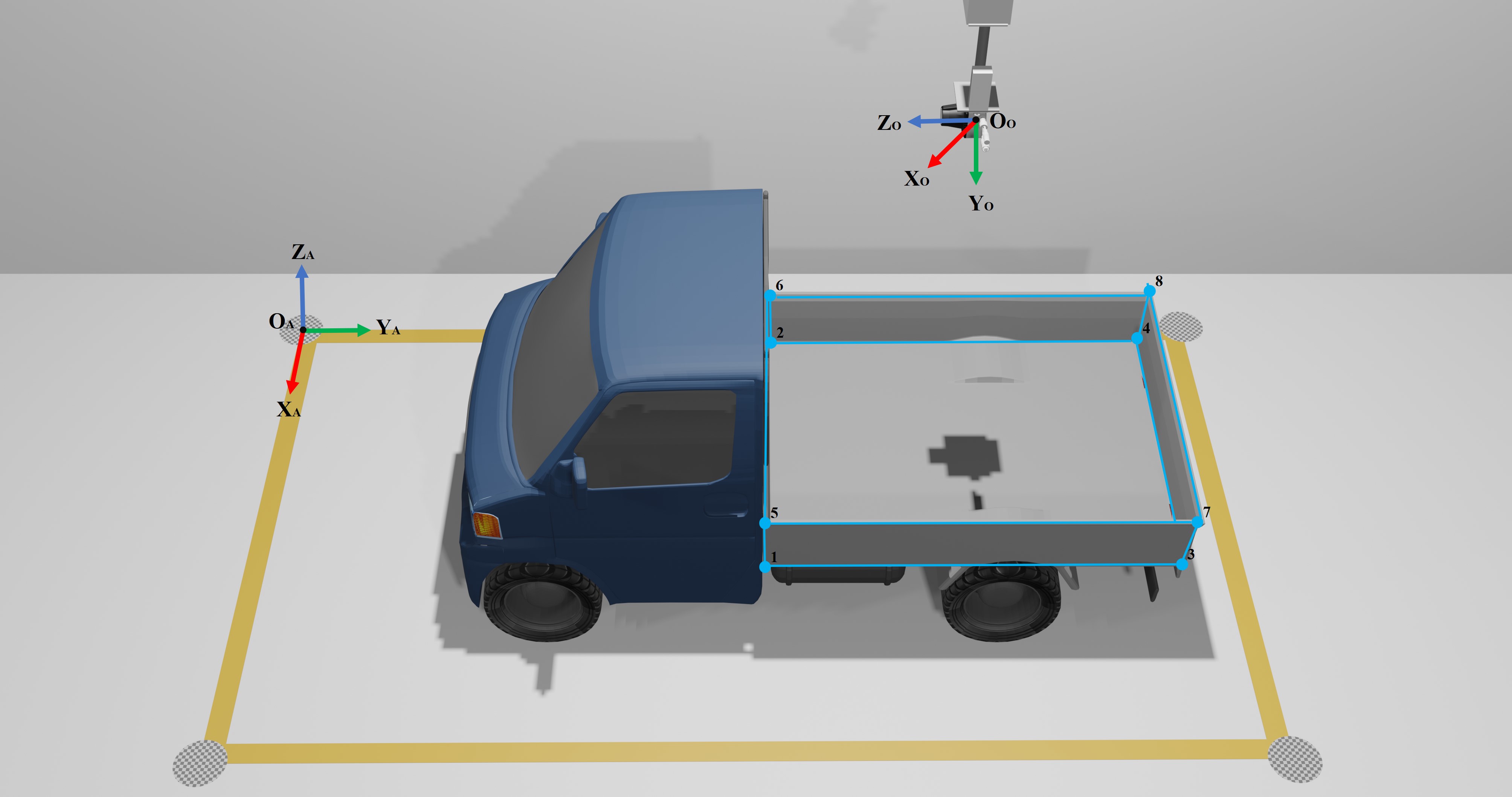}
        }
	\caption{Establish world coordinate system. (a) Installation location. Four reflective boards are installed at the corners of the parking area, with red dots indicating the center of each reflective board. (b) World coordinate system. The global coordinate system $(O)$ is established based on these reflective boards. Blue dots represent the compartment corner points, and the blue area denotes the loading area, all indicated under coordinate system $(A)$.}
	\label{software1-2}
\end{figure}

\subsection{Establish World Coordinate System and Segment Vehicle Point Cloud}
As illustrated in Fig. \ref{software1-2}(a), to provide usable coordinates for the mobile manipulator, our system defines a unified world coordinate system $(A)$, to which the LiDAR-acquired point cloud is transformed. Additionally, we segment the vehicle point cloud from the scene to improve detection speed and accuracy.

We propose a localization method based on round reflective boards that can simultaneously establish the world coordinate system $(A)$ and segment vehicle point clouds. As shown in Fig. \ref{software1-2}(b), reflective boards are placed at the four corner points of the pre-planned parking area. The specific steps are divided into coarse positioning and fine positioning. Considering that high-reflectivity objects may also exist in the actual loading scene, we do not recommend directly filtering out low-reflectivity points to obtain the point cloud of the reflective boards. A better approach for coarse positioning is to manually select the region containing each reflective board and then filter out low reflectivity points based on a reflectivity threshold. Subsequently, the random sampling consistency algorithm (RANSAC) is used to estimate the center of each reflective board point cloud to obtain accurate positioning results.
\begin{figure}[!t]
	\centering 	
        \subfigure{
		\includegraphics[width=3.2in]{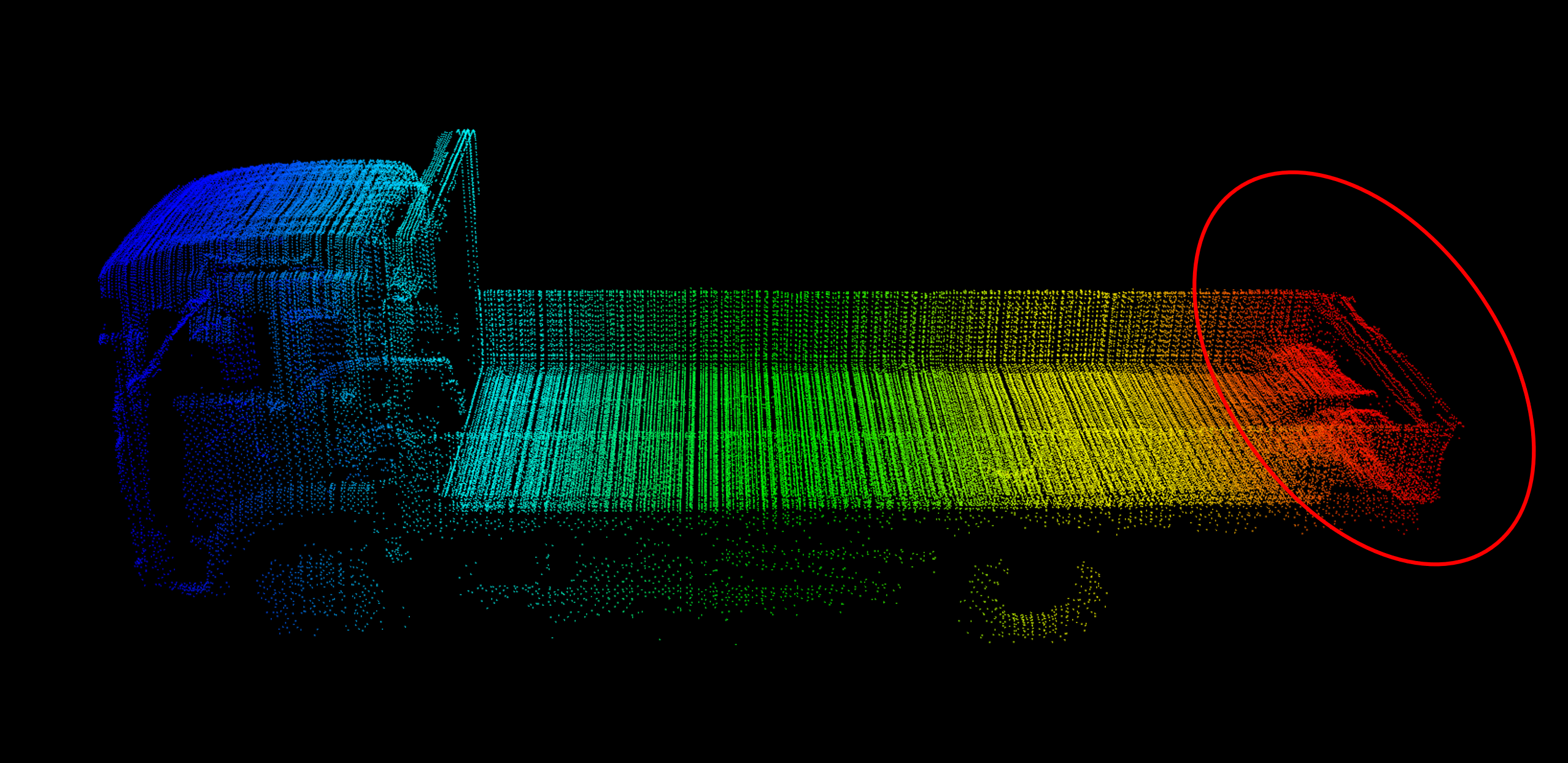}
	}
	\caption{Visualization of public dataset vehicles. The red box area shows that the vehicles have occlusions and missing point clouds at the rear.}
	\label{software_5}
\end{figure}
If the installation position of the LiDAR and the planned parking area remain unchanged, it only needs to be executed once. The world coordinate system $(A)$ is located in front of the right side of the vehicle parking position, with the corner point ($P_{1}$) of the rectangular area as the origin. The directions of the length and width of the rectangle serve as the x-axis and y-axis, respectively, and the normal vector of the plane serves as the z-axis. The specific calculation method is as follows:
\begin{align}
    &{P_{AORG}}={P_{1}}, \notag \\
    &\overline{OX_{\text {A}}}=\frac{\overline{P_{2}P_{1}}}{|\overline{P_{2}P_{1}}|}, \notag \\
    &\overline{OZ_{\text {A}}}=\frac{\overline{P_{3}P_{1}}}{|\overline{P_{3}P_{1}}|}, \notag \\
    &\overline{O{Y}_{A}}=\overline{O{X}_{A} }\times\overline{O{Z}_{A} }.
\end{align}

The world coordinate system $(A)$ can be described in the rotating platform coordinate system $(O)$ as follows:
\begin{equation}
    {^{O}_{A}T}=
    \begin{bmatrix}
    {^{O}_{A}R}&{P_{AORG}}\\ 
    0 &1 
    \end{bmatrix},
\end{equation}
where, ${^{O}_{A}R}=[\overline{OX_{A}}\enspace \overline{OY_{A}}\enspace \overline{OZ_{A}}]$. Further, the description of the rotating platform coordinate system $(O)$ under the world coordinate system $(A)$ can be obtained:
\begin{equation}
    {^{A}_{O}T}={
    \begin{bmatrix}
    {^{O}_{A}R}&{^{O}P_{AORG}}\\ 
    0 &1 
    \end{bmatrix}
    }^{T}.
\end{equation}

According to the homogeneous transformation matrix $^{A}_{O}T$, the point cloud in the rotating platform coordinate system ($O$) is transformed into the world coordinate system ($A$). By constraining each point within specified ranges along the three axes, a clear point cloud of the vehicle compartment is obtained.
\begin{align}
    ^{A}{P}_i &= {_{O}^{A}T}  \hspace{2pt} {^{O}P_{i}}, \notag \\ 
    M_P &= \{ (x, y, z) \in \mathbb{R}^3 \mid \begin{aligned}
           & 0 \leq x \leq X_{\text{max}}, \\
           & 0 \leq y \leq Y_{\text{max}}, \\ 
           & Z_{\text{min}} \leq z \leq Z_{\text{max}} \}.
         \end{aligned}
\end{align}
where $X_{max}={\mid{^{A}P_{2}}{^{A}P_{AORG}}\mid}$, $Y_{max}={\mid{^{A}P_{4}}{^{A}P_{1}}\mid}$, and $M_{P}$ represents the set of points in the parking area. To filter the point cloud on the ground and limit the height range of the detection area, $Z_{min}$ is set at a distance of 30 cm from the ground height, and $Z_{max}$ is set at a distance of 30 cm from the bottom of the LiDAR.

\begin{figure}[!t]\color{black}
	\centering 	
        \subfigure{
		\includegraphics[width=3.4in]{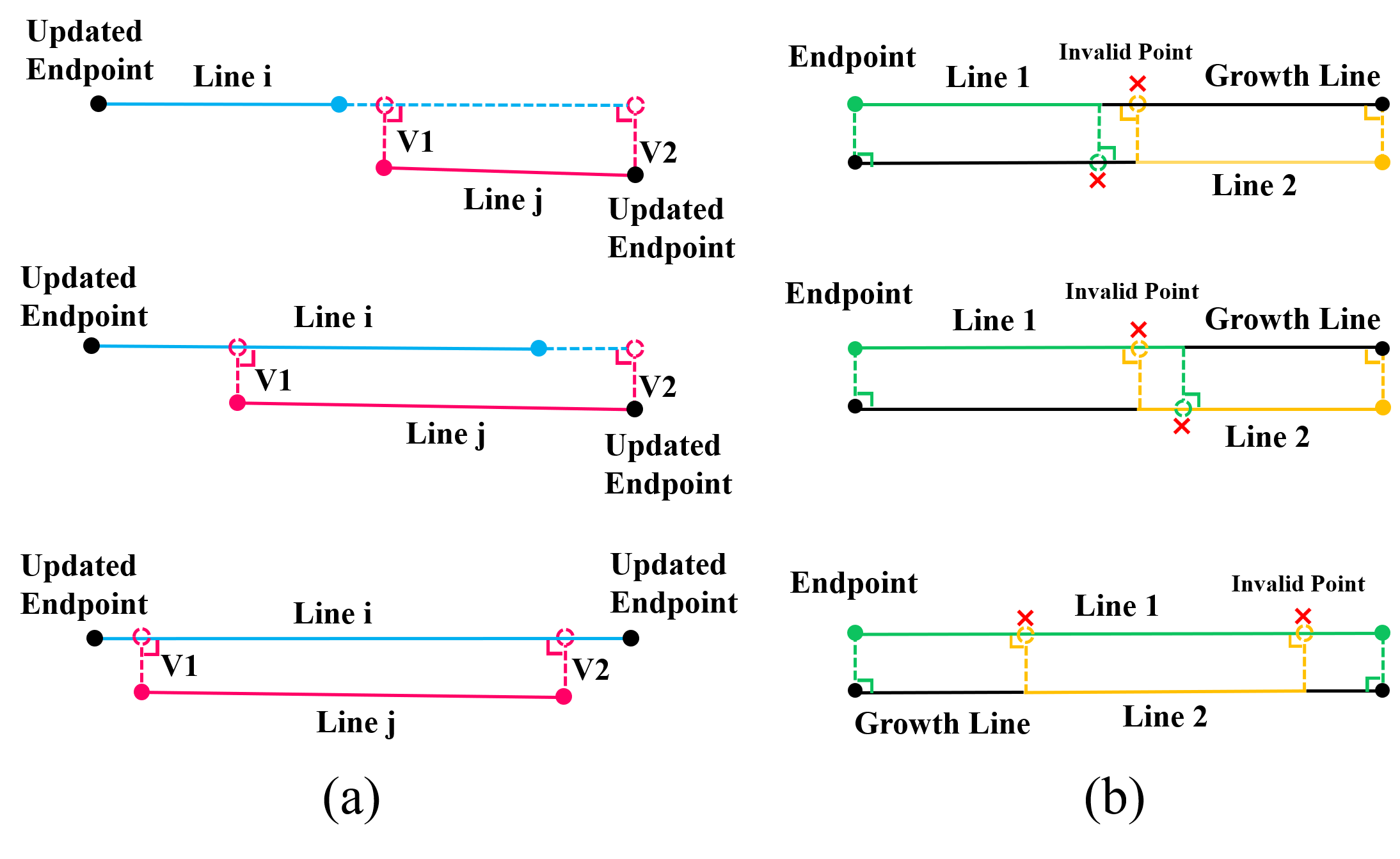}
	}
	\caption{Illustration of the contour fused and completion strategy. (a) Three cases of line segment fusion within the plane. The blue and red segments represent two line segments that satisfy the merging criteria, while the black points are the endpoints of the new line segment. \textcolor{black}{(b) Three cases of truck compartment contour line completion. The green and yellow segments represent the compartment edge segments, and the black lines and points indicate the newly grown line segments and endpoints, respectively.}}
	\label{merge_grow}
\end{figure}

\begin{figure}[!t]\color{black}
	\centering 	
        \subfigure{
		\includegraphics[width=3.0in]{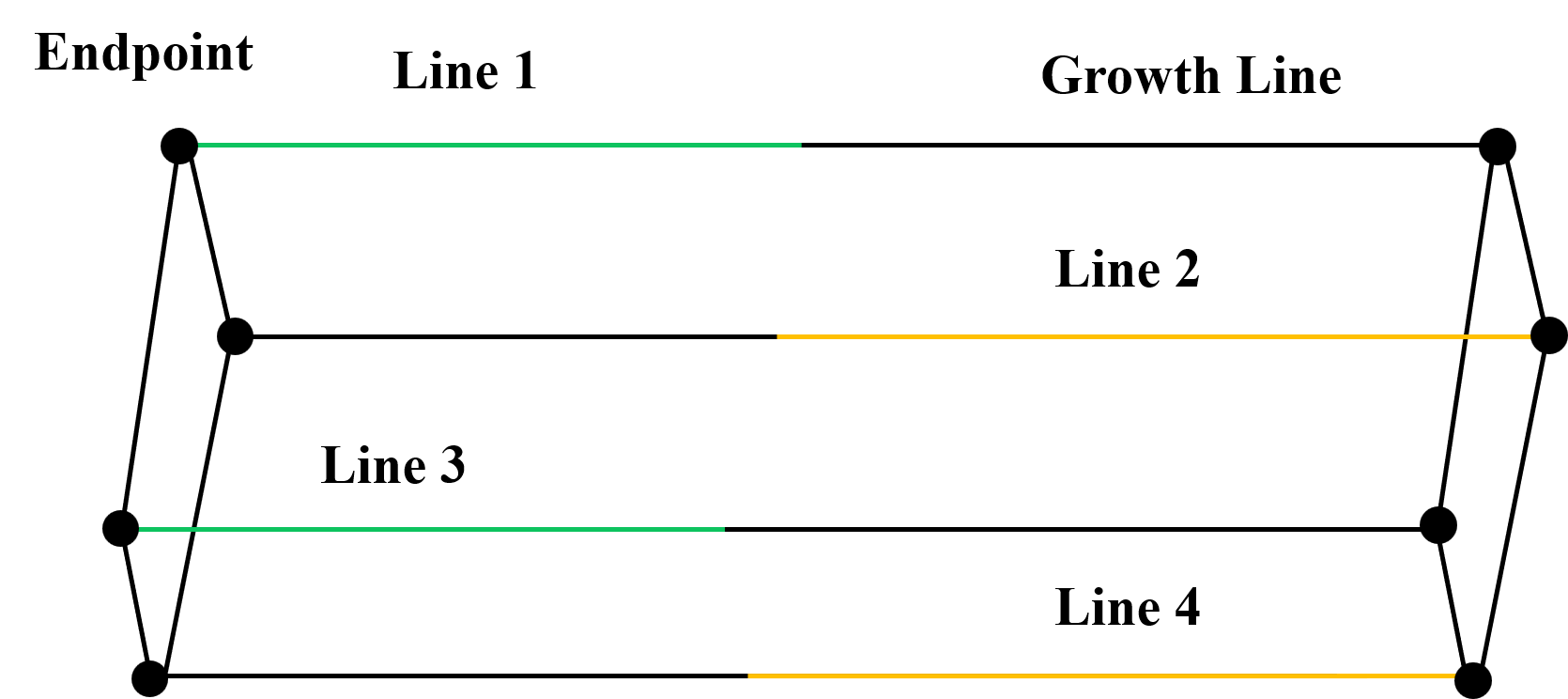}
	}
	\caption{The result of contour line completion. The blue lines are the segments to be completed, while the black lines are the new segments. The black endpoints represent the eight key points of the truck compartment.}
	\label{grow_res}
\end{figure}

\subsection{Key Point Localization of Compartment}
The complete geometric contour of the truck compartment is crucial for the subsequent localization of key points. As shown in Fig. \ref{software_5}, some fenced truck compartments lack rear fences, resulting in point cloud data missing rear plane information and containing foreign objects inside the compartment. This leads to issues such as missing or broken 3-D line segments on the compartment surface, affecting the determination of line segment endpoints. Therefore, it is necessary to further optimize the extracted contour line features. Additionally, we found that the method \cite{ref20} is robust in detecting the longest edges on the sides and bottom of the carriage, but less effective for other directional contour lines. This paper investigates a strategy for in-plane contour line fusion and completion. \textcolor{black}{Unlike previous methods that perform line fusion across the entire 3-D space, our approach first fuses line segments within each plane, then performs line segment clustering across the entire space to reconstruct and complete the fractured and missing 3-D line segments along the compartment edges.}

\begin{algorithm}[!t]
\caption{Key Point Localization of Compartment}
\renewcommand{\algorithmicrequire}{\textbf{Input:}}
\renewcommand{\algorithmicensure}{\textbf{Output:}}
\begin{algorithmic}[1]
\REQUIRE $S = \{l_1, \ldots, l_i\}$: list of 3-D line segments, $Th_a$, $Th_b$.
\ENSURE $\hat{P} = \{p_1, \ldots, p_j\}$: key points of compartment.
\STATE $G$; // Initialized as group list
\STATE $C$; // Initialized as cluster line segments
\STATE $\hat{P}$; // Initialized as unknown

\FOR{lines in planes}
    \STATE LineMerging(lines);
\ENDFOR
\STATE $S = S'$ // Update 

\STATE Search for the longest line Lmax;

\FOR{$i = 1$ to $n$}
    \STATE Compute cosine similarity $C_i$ according to Eqs. (15);
    \STATE Compute distance $D_i$: Distance($L_i$, $L_1$);
    \IF{$C_i \leq Th_a$ \AND $D_i \leq Th_b$}
        \STATE Add $L_i$ to $C$;
    \ENDIF
\ENDFOR

\STATE Select the top four segments from $C$ as $L_{\text{max}}, L_2, L_3, L_4$;
\STATE $G = $ LinesGrouping($L_{\text{max}}, L_2, L_3, L_4$);

\STATE Add the endpoint of $L_{\text{max}}$ to $\hat{P}$;
\FOR{each line $L_i$ in $\{L_{\text{max}}, L_2, L_3, L_4\}$}
    \FOR{each endpoint $p$ of $L_i$}
        \STATE Compute projection vector $P_{roj}$ according to Eqs. \ref{con:cossim};
        \STATE Compute projection point $q$;
        \IF{$q$ is valid point}
            \STATE Add $q$ to $\hat{P}$;
        \ELSE
            \STATE Add $p$ to $\hat{P}$;
        \ENDIF
        
    \ENDFOR
\ENDFOR

\RETURN key points of compartment $\hat{P} = \{p_1, \ldots, p_j\}$;
\end{algorithmic}
\end{algorithm}

\subsubsection{\textcolor{black}{Spatial Line Segment Fusion}}
\textcolor{black}{First, for each 3-D line segment, the latitude of each internal point is calculated as $arcsin(z)$, where $z$ is the component of the normalized direction of the line. The 3-D line segments are grouped into intervals with a latitude step size of 6 degrees. Starting with the longest 3-D line segment $L_{i}$, other line segments $L_{j}$ within the same latitude interval and plane are examined. The distances from the origin to the start points of $L_{i}$ and $L_{j}$ are calculated as $d_{L_{i}}$ and $d_{L_{j}}$, respectively, as well as the distance $d_{mag}$ between the first point of the input point cloud and the origin. If $\left|d_{\mathbf{L}_i}-d_{\mathbf{L}_j}\right| / d_{mag} < 0.1$, and the perpendicular distances $V_1$ and $V_2$ from the two endpoints of $L_j$ to $L_i$ are both less than $4s_k$ (where $s_k$ is the scale of the 3-D plane), these line segments are fused. As shown in Fig. \ref{merge_grow}(a), the updated line segment endpoints are the two farthest points among the four endpoints of $L_{i}$ and $L_{j}$.}

\subsubsection{Spatial Line Segment Clustering} 
First, determine the clustering center, then assign each line segment to the clustering center based on their length and directional characteristics. Finally, sort the line segments by length within each cluster and sequentially select the top four longest segments as the compartment edge contour lines. The process is as follows: calculate the length of all line segments, and select the longest one as the reference edge, the clustering center. Excluding the reference edge, sort each line segment by length, and sequentially compute the cosine similarity between the line segments and the reference edge. 
\begin{equation}
C_i\left(l_i, l_{\max }\right)=\frac{\overline{l_il_{max}}}{|\overline{l_il_{max}}|}.\label{con:cossim}
\end{equation}

Among the segments that meet the threshold, find the three longest ones as the contour lines of the carriage, denoted as $l_{2}$, $l_{3}$, and $l_{4}$. The four contour lines are classified into two groups based on their distances from $l_{1}$. The line closest to $l_{1}$ is designated as $l_{2}$, thus $l_{1}$ and $l_{2}$ form one group, while $l_{3}$ and $l_{4}$ form the other group. Additionally, it is necessary to determine the positional relationship between $l_{1}$, $l_{2}$, and $l_{3}$, $l_{4}$. This is done by calculating the cosine similarity $S$ between the projection vector from the endpoint of $l_{2}$ to $l_{1}$ and the projection vector from the endpoint of $l_{3}$ to $l_{4}$. If $S > 0$, then $l_{1}$ and $l_{4}$ are at the same horizontal position. Otherwise, the positions of $l_{3}$ and $l_{4}$ are swapped.
\begin{equation}
Proj=\overline{P_{1}P_{e}}-(\overline{P_{1}P_{e}} \cdot \frac{\overline{P_{1}P_{e}}}{|\overline{P_2P_1}|}){|\overline{P_2P_1}|},
\end{equation}
where $P_{e}$ is the given projection point, and $P_{1}$ and $P_{2}$ are the endpoints of the projection line.

\subsubsection{Contour Line Completion}
\textcolor{black}{Given the clustered edge segments of the truck compartment, the method performs the following steps to complete the contour lines. As shown in Fig. \ref{merge_grow}(b), growth starts from any endpoint of the reference edge $l_{max}$ toward $l_{2}$. The projection direction of the endpoint of $l_{max}$ onto $l_2$ determines the growth direction, and the projection point on $l_2$ serves as the termination point for the growth. If the projection point of $l_{max}$ lies within $l_2$, it is deemed an invalid point. Otherwise, the point farthest from the endpoint of $l_2$ is defined as the termination point. The complete contour line is obtained by sequentially completing the clustered edge segments of the compartment, as illustrated in Fig. \ref{grow_res}. The vertices of the completed contour line are defined as the eight key points of the truck compartment. Algorithm 1 summarizes the process for locating these key points.}

\begin{figure}[!t]\color{black}
	\centering 	
        \subfigure[]{
		\includegraphics[width=1.5in]{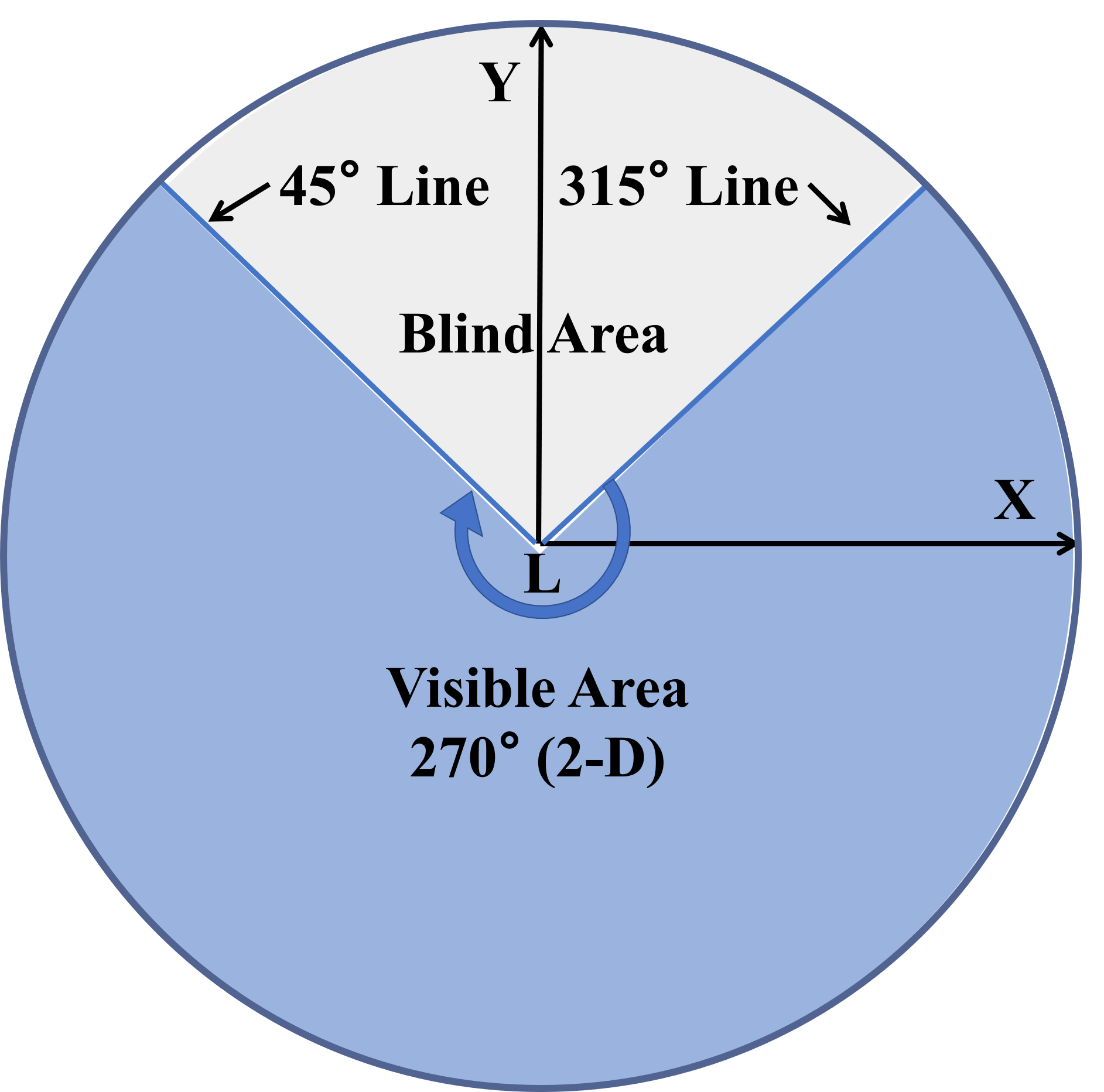}
	}\hspace{-0.8em}
        \subfigure[]{
            \includegraphics[width=1.5in]{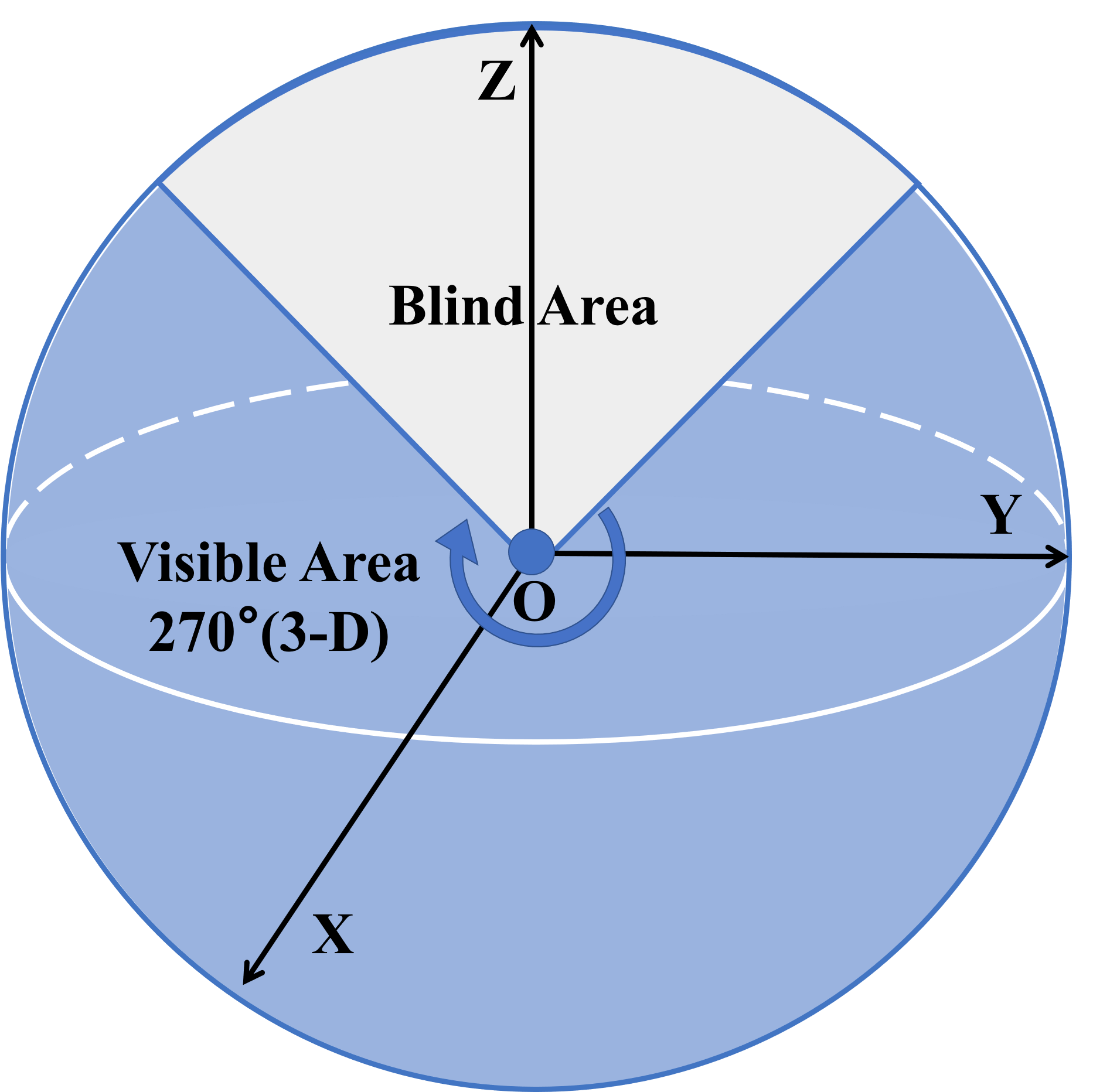}
        }
	\caption{LiDAR field of view. (a) 2-D LiDAR field of view. \textcolor{black}{The 2-D LiDAR scans from the starting scan line (45\degree) to the ending scan line (315\degree), with a horizontal field of view of 270\degree.} (b) Wide field-of-view 3-D LiDAR. \textcolor{black}{The rotating platform moves the 2-D LiDAR, extending its field of view to cover a spherical space in the range of 180\degree-270\degree, with both the horizontal and vertical fields of view exceeding 180\degree.}}
	\label{hardware_2}
\end{figure}

\begin{figure}[!t]
	\centering 	
        \subfigure{
		\includegraphics[width=3.2in]{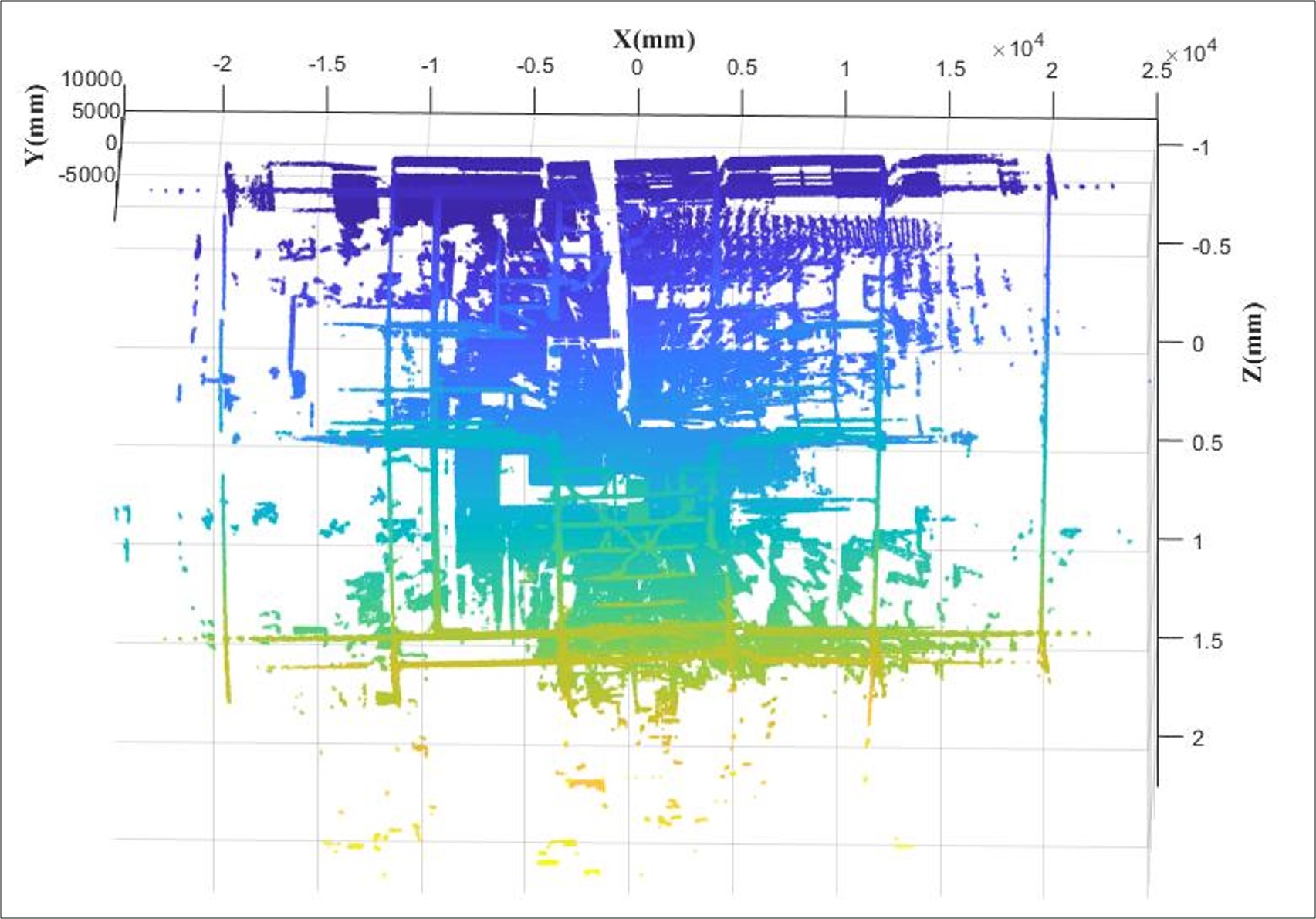}
	}
	\caption{Scene point cloud acquired by LiDAR. The LiDAR is fixed 3.3 \,m above the ground and collects point clouds while rotating 180\degree.}
	\label{wildfield}
\end{figure}

\section{Experiments}
\subsection{Experimental Setup}
\subsubsection{Prototype System}
The sensors utilized in this system include the Rich Lakibeam 1 and the Kinco SMC40S-0005-30Q8K-5DSU. The Rich Lakibeam 1 is a 2-D LiDAR used as the primary ranging device of the system, featuring a plane scanning range of $270\degree$, an angular resolution of $0.25\degree$, and a scanning frequency of \textcolor{black}{$25$ Hz}, as shown in Fig. \ref{hardware_2}(a). The Kinco SMC40S-0005-30Q8K-5DSU, mounted at the bottom of the LiDAR, enables rotational scanning movement for the 2-D LiDAR. The entire device provides a field of view between $180\degree$ and $270\degree$ in three-dimensional space, as shown in Fig. \ref{hardware_2}(b). In practical measurements, the LiDAR is installed at a height of 3.3 \text{m}. For system control, a SIMATIC S7-1200 controller is employed, and the Kinco FD114S-CB-006 serves as the servo motor driver. Data processing is conducted on a PC equipped with a 6-core Intel i7 CPU (\textcolor{black}{2.60 GHz}). The point cloud scanned by the LiDAR in the factory is shown in Fig. \ref{wildfield}. Compared to conventional LiDAR systems, our device offers a significantly broader field of view.

\subsubsection{Dataset}
To further verify the performance of the localization algorithms, we conducted additional experiments on the public dataset ZPVehicles\cite{ref23}. ZPVehicles is a large vehicle high-density point cloud dataset collected by LiDAR, comprising high-density point cloud data of approximately 800 vehicles, including box trucks, warehouse trucks, fence trucks, cranes, semi-trailer tractors, buses, minibusses, garbage trucks, water sprinklers, communication storage vehicles, and tank trucks. The fence truck model consists of 106 large vehicle point clouds of varying shapes, with truck lengths ranging from 5 to 13 meters and containing internal obstructions. Due to the characteristics and installation constraints of the data collection equipment, the point cloud data for each vehicle in the dataset lacks front and rear plane information, presenting a certain challenge. To calculate the error, we annotated the compartments of two vehicle types, fence trucks and vans.

\subsubsection{Implementation Details}
The scanning frequency of the 2-D LiDAR is set to 10 Hz, with a scanning range configured between $45\degree$ and $315\degree$. The resolution of the servo motor is set to $0.2\degree$, and its scanning range is from $0\degree$ to $180\degree$. For cropping of the region of interest, ${Z}_{min}$ is set to $0.3$ m, and ${Z}_{max}$ to 4.0 m, which is suitable for most vehicles. Additionally, we provide general parameter settings: $\textcolor{black}{k=33}$, ${\textcolor{black}{\delta=19\degree}}$ , ${th}_{O}={S}_{P_{S}}$, ${th}_{P}=50{S}_{P_{S}}$, where ${S}_{P_{S}}$ is the scale of point ${P}_{S}$.
\subsubsection{Metrics}
To evaluate the accuracy of key point localization and the effectiveness of the predicted area, we use an average distance metric to evaluate the performance of the compartment key point localization algorithm. This metric calculates the average distance between the estimated eight key points $(p_i)$ and the ground truth $(q_i)$.
\begin{equation}
    ADD=\frac{1}{m}\displaystyle\sum_{i=1}^{m}{\left \| {p}_{i}-{q}_{i}\right \|}_{2}.
\end{equation}

For an individual truck, the estimated key points are considered correct if the average distance is less than 5\%, 7\%, or 10\% of truck dimensions. The following formula represents the calculation method for a 5\% threshold:
\begin{equation}
Th_{e}=\frac{(l+w+h)}{3} \cdot 5 \%,
\end{equation}
where $l$, $w$, and $h$ are the length, width, and height of the truck, respectively.

\begin{figure}[!t]
	\centering 	
        \subfigure{
		\includegraphics[width=3.4in]{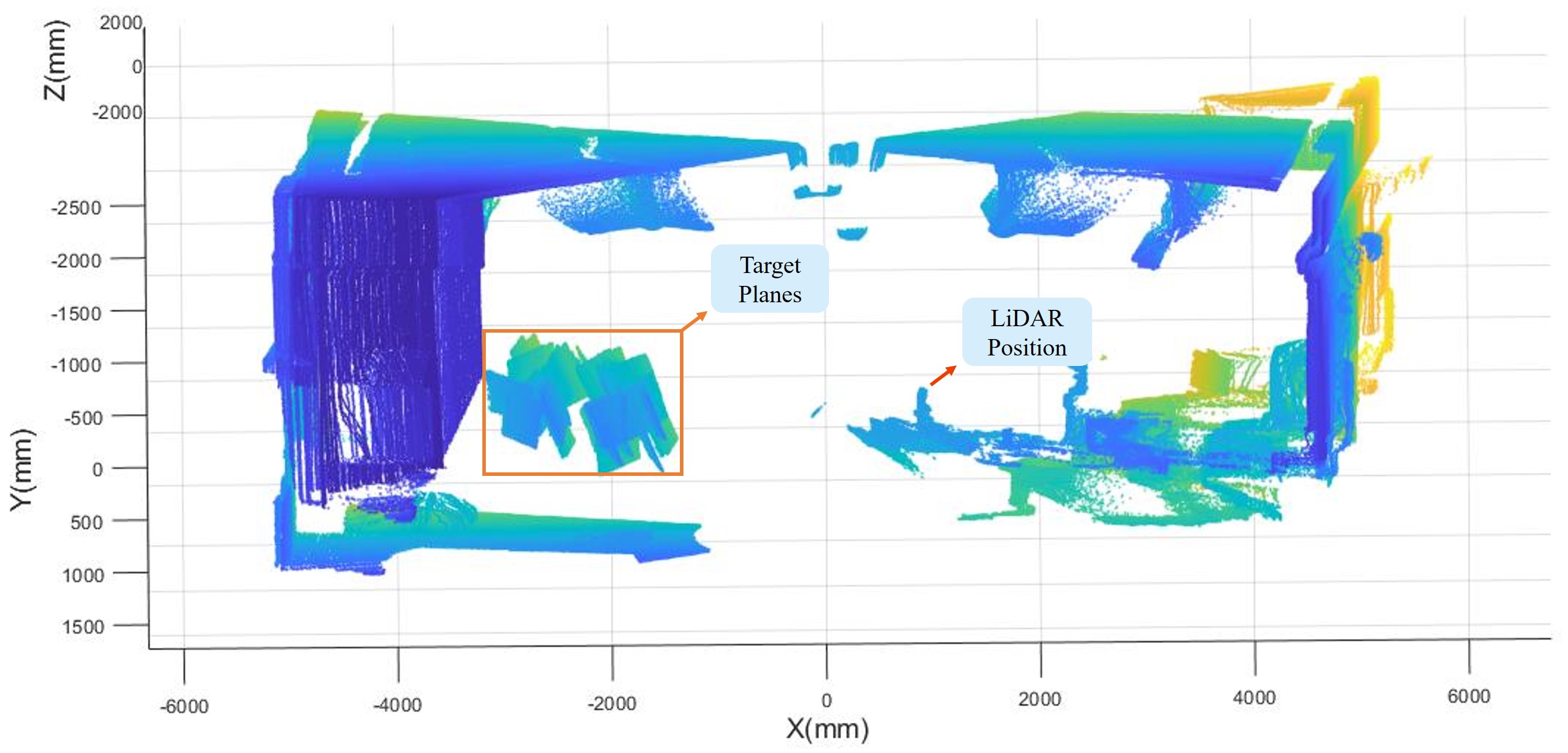}
	}
	\caption{Position of the target planes. The yellow boxes indicate 30 target planes. For each plane, the rotating platform scanned the target plane twice with 360-degree rotations to obtain a data set.}
	\label{calib}
\end{figure}

\begin{figure*}[!t]
	\centering 	
        \subfigure{
		\includegraphics[width=6.8in]{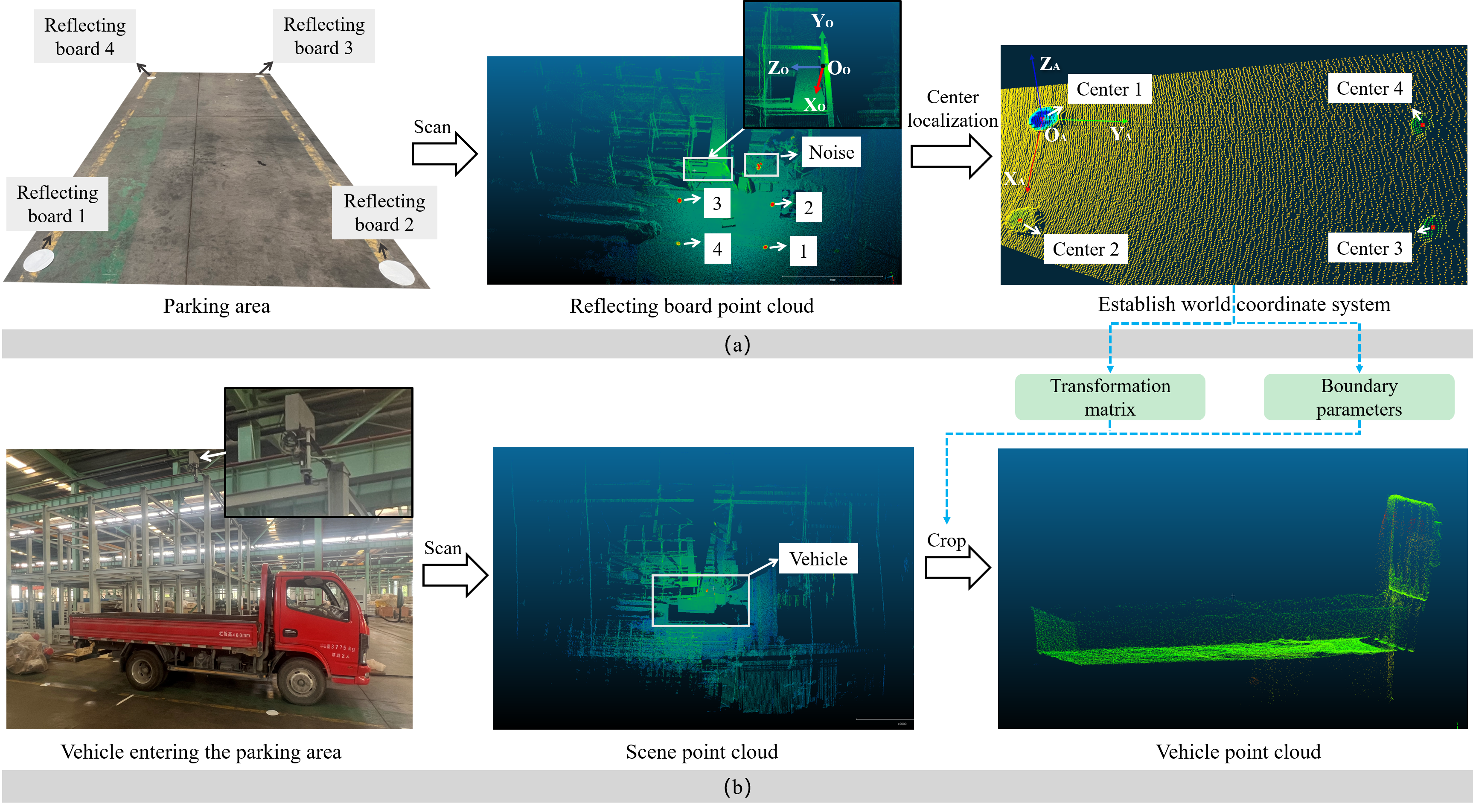}
	}
	\caption{\textcolor{black}{Establishing world coordinate system and segmentation. (a) Parking area localization. (b) Vehicle point cloud segmentation. Once the device is installed, it performs a one-time localization of the parking area. Yellow lines are parking areas, and white are round reflective boards. Extracting the point cloud within the region of interest improves the efficiency of subsequent detection. The world coordinate system provides unified 3-D coordinates for the mobile manipulator.}}
	\label{software_6}
\end{figure*}

\begin{figure}[!t]\color{black}
	\centering 	
        \subfigure[]{
		\includegraphics[width=1.6in]{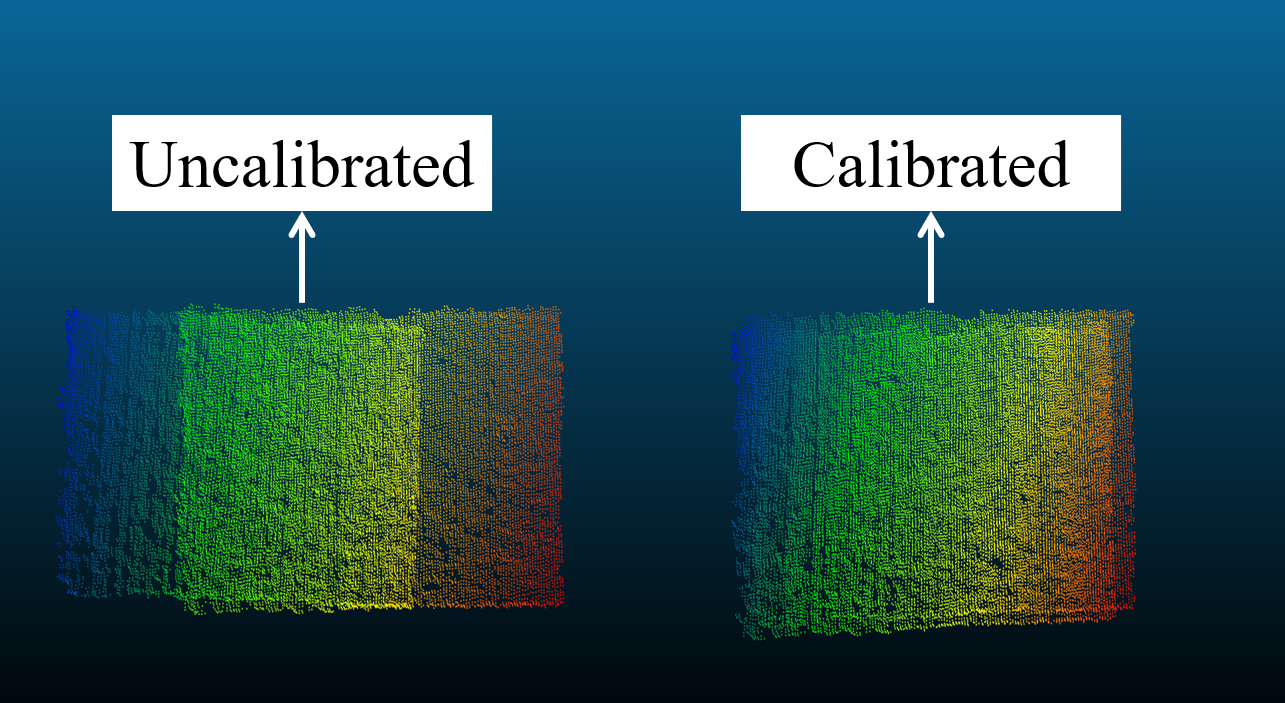}
	}\hspace{-0.7em}
        \subfigure[]{
            \includegraphics[width=1.6in]{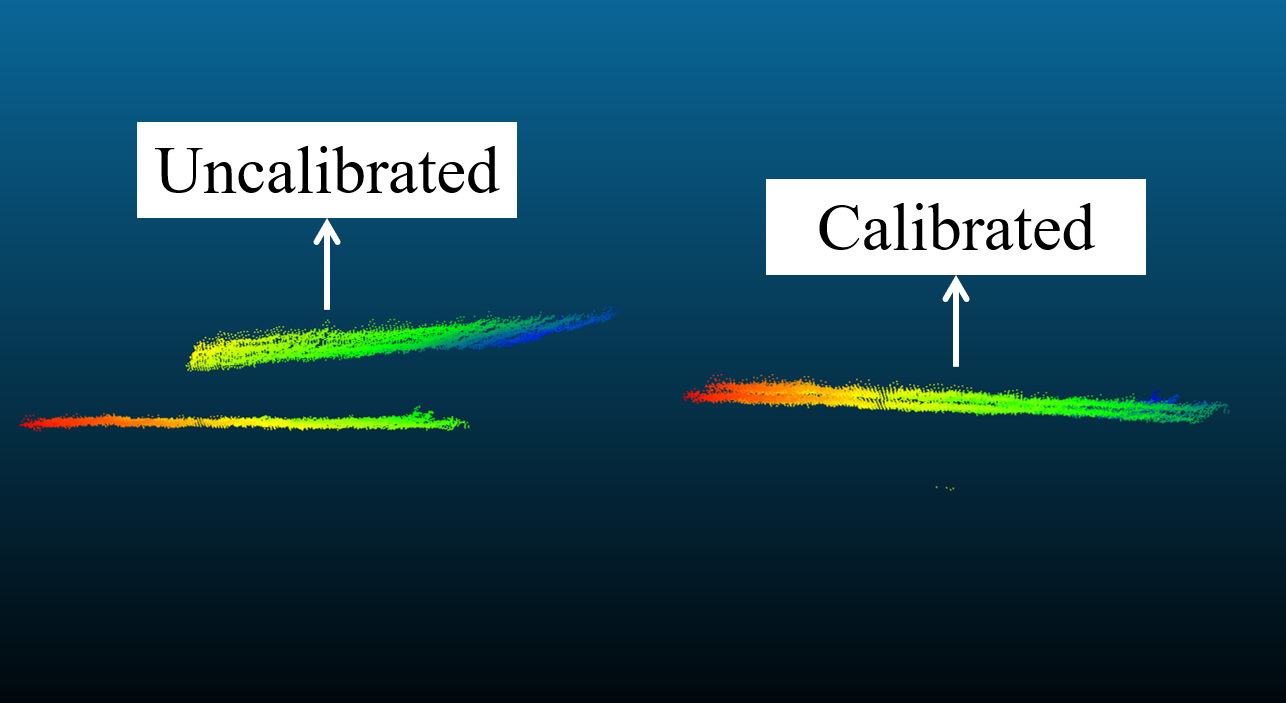}
        }\hspace{-0.7em}
        \subfigure[]{
            \includegraphics[width=1.6in]{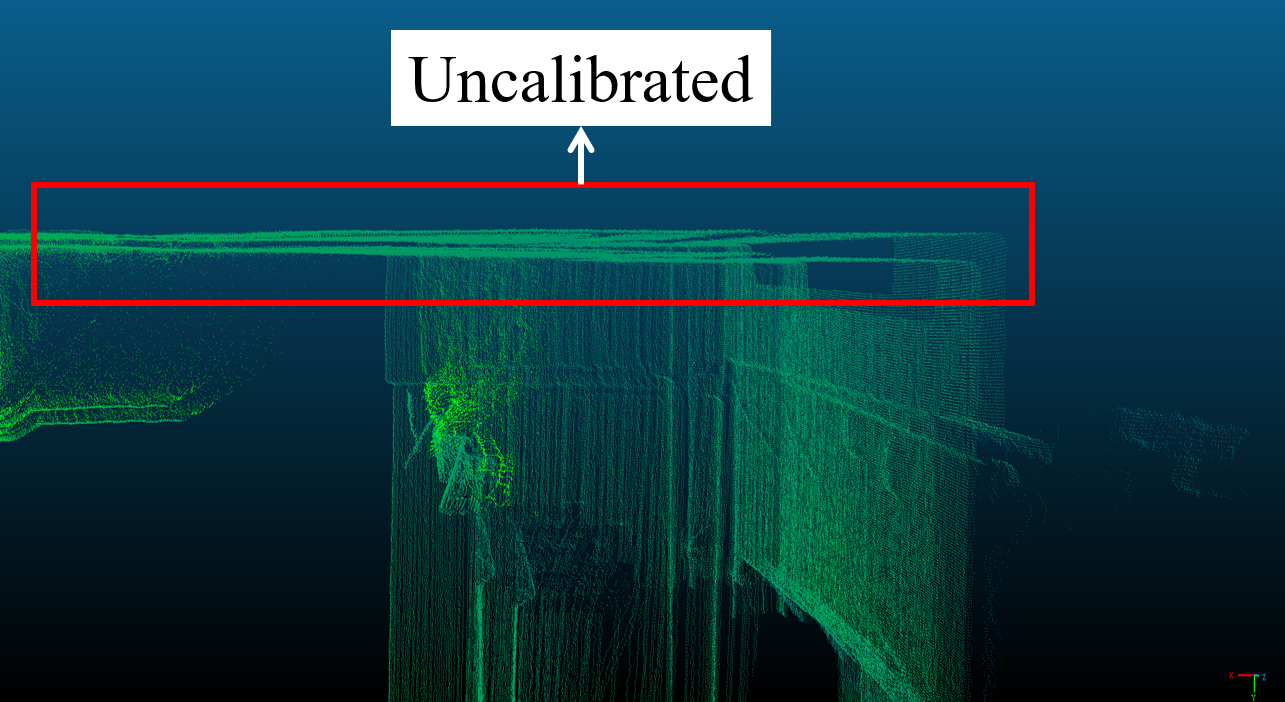}
        }\hspace{-0.7em}
        \subfigure[]{
            \includegraphics[width=1.6in]{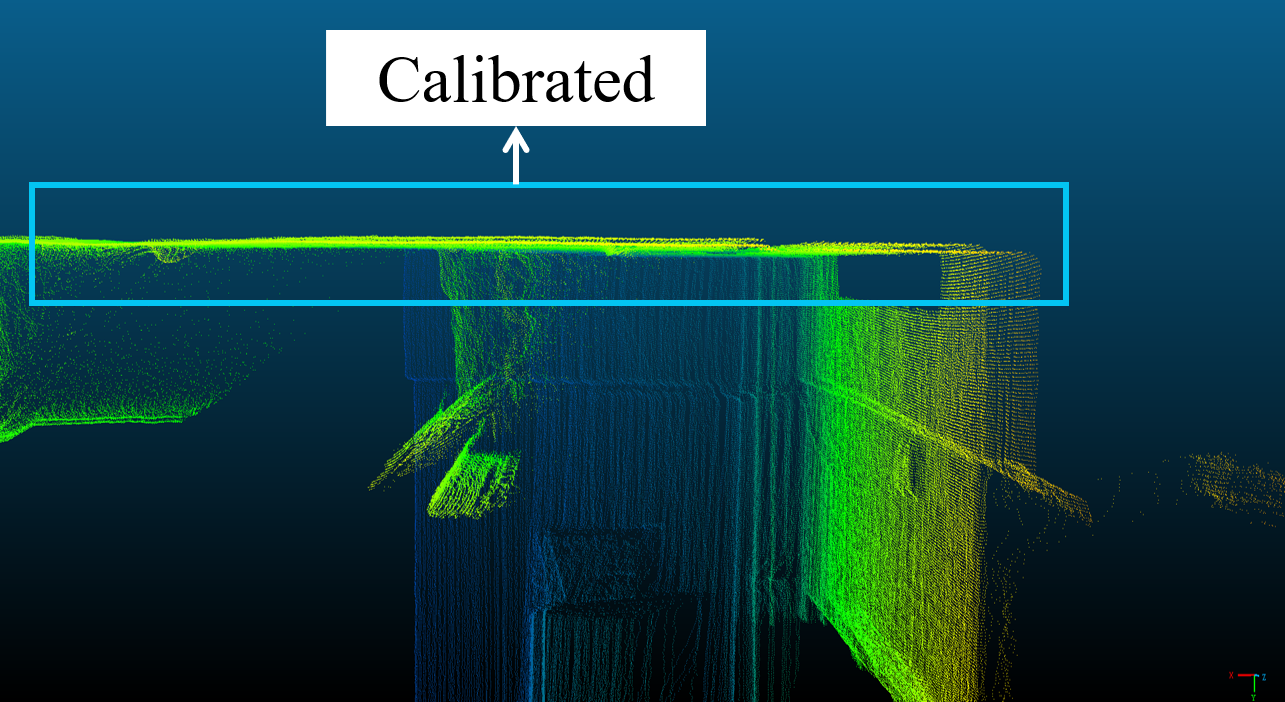}
        }
	\caption{\textcolor{black}{Comparison of calibrated and uncalibrated planes. (a) Front view of calibration board. (b) Top view of calibration board. (c) Uncalibrated walls. (d) Calibrated walls.}}
	\label{ca_v}
\end{figure}

\subsection{Calibration of Rotating Platforms}
To minimize the drift effect of the 2-D LiDAR, data collection starts at least one hour after activating the LiDAR. We calibrated 30 target planes, as shown in Fig. \ref{calib}(yellow boxes). The normal vector $n$ of the target plane is far away from the degenerate case \( n = [0\;\pm1\;0] \), and the horizontal distance between the target plane and the LiDAR is kept between 1.6 and 3.2 meters. \textcolor{black}{A comparison of a single calibration plane from different perspectives before and after calibration is shown in Fig. \ref{ca_v}(a) and \ref{ca_v}(b). For a single calibration plane, the rotating platform performed a $360\degree$ rotation, generating two planar point clouds. Due to assembly errors, the two-point clouds did not initially overlap. After calibration, the degree of overlap between the point clouds increased, indicating that the calibration was effective. The same calibration parameters were applied to the wall point cloud, significantly improving the point cloud quality.} The final calibration results are presented in Table \ref{tab:calib}, the 2-D LiDAR is positioned relatively far from the rotation axis, resulting in a significant value for the z-axis calibration parameter. Due to the degeneracy in single-axis rotation\cite{ref21}, the rotation and translation degrees of freedom along the rotation axis (y-axis) could not accurately calibrate.

\begin{table}[!t]%
	\centering
	\renewcommand{\arraystretch}{1.2}
    \setlength{\tabcolsep}{12pt}
	\caption{Calibration Results of Rotating Platform}
	\footnotesize
	\begin{tabular}{cc}
		\toprule
		\textbf{Calibration parameter}   &\textbf{Value} \\
		\midrule
		$w_x$(degrees)  & -1.46  \\ 
		$w_y$(degrees)  & 5.36  \\ 
		$w_z$(degrees)  & -0.45  \\
		$t_x$(mm)   & 14.56  \\
		$t_y$(mm)   & -0.16  \\
		$t_z$(mm)   & 119.67  \\
		\bottomrule
	\end{tabular}%
	\label{tab:calib}%
\end{table}%

\subsection{Experiment of Establishing World Coordinate System and Locating Parking Area}
As shown in Fig. \ref{software_6}, round reflective boards are pasted at the four intersections of the ground rectangular parking area (yellow line). The LiDAR rotates $180\degree$ around the rotation axis from the initial position to measure the point cloud in the scene, including the point cloud of vehicles and other objects. In the scene, objects with higher reflectivity have greater brightness in the point cloud, as shown in the red and yellow areas in Fig. \ref{software_6}. In addition to the four reflector point clouds, there are other highly reflective objects in the scene. Use the method proposed in section \uppercase\expandafter{\romannumeral3} to extract the reflector center and establish the world coordinate system $(A)$. The establishment result of the coordinate system is shown in Fig. \ref{software_6}(a), and we used transformation matrices and boundary parameters to crop the point cloud of the truck compartment, as shown in Fig. \ref{software_6}(b).

\subsection{\textcolor{black}{Comparison with State-of-the-Art Methods}}
\begin{figure*}[!t]\color{black}
	\centering 	
        \subfigure[]{
		\includegraphics[width=1.3in]{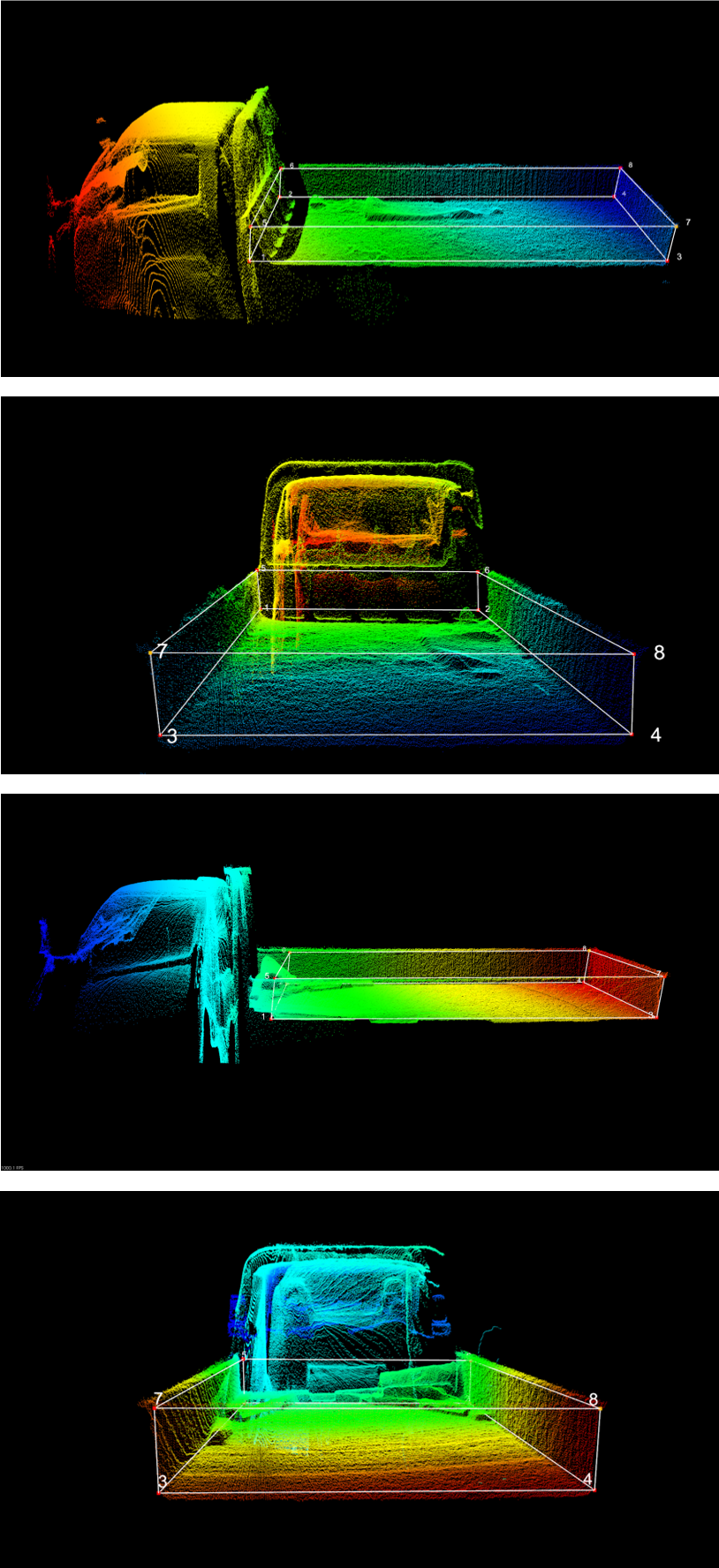}
	}\hspace{-0.7em}
        \subfigure[]{
            \includegraphics[width=1.3in]{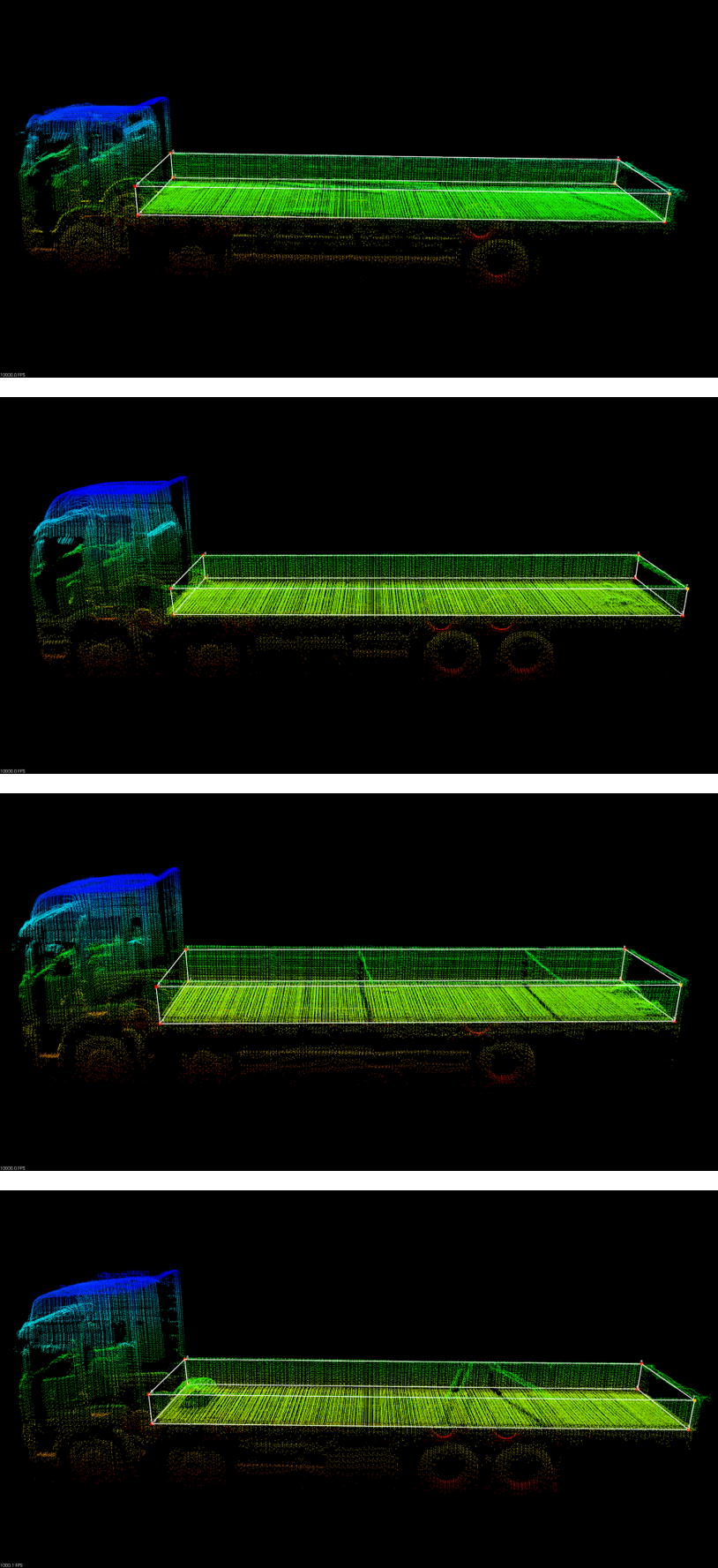}
        }\hspace{-0.7em}
        \subfigure[]{
            \includegraphics[width=1.3in]{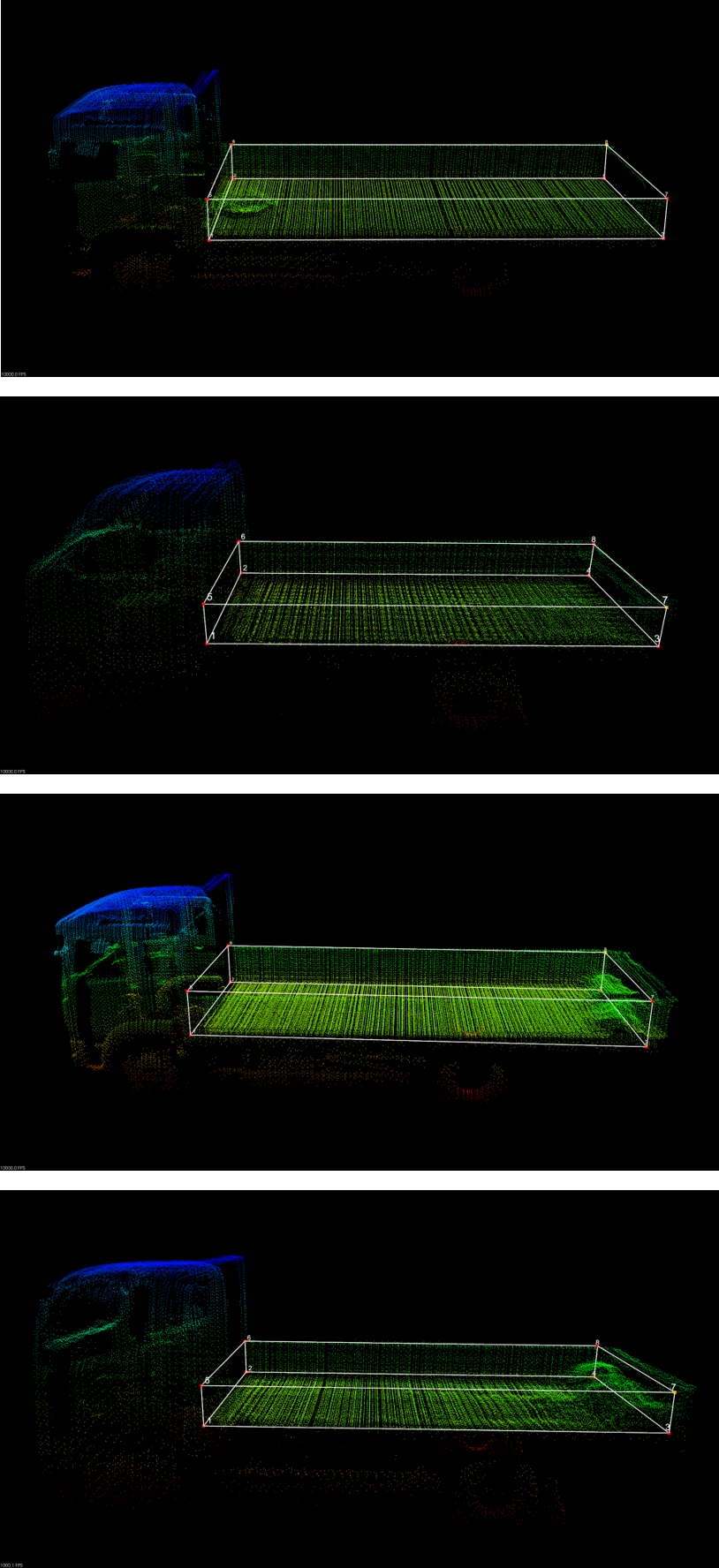}
        }\hspace{-0.7em}
        \subfigure[]{
            \includegraphics[width=1.3in]{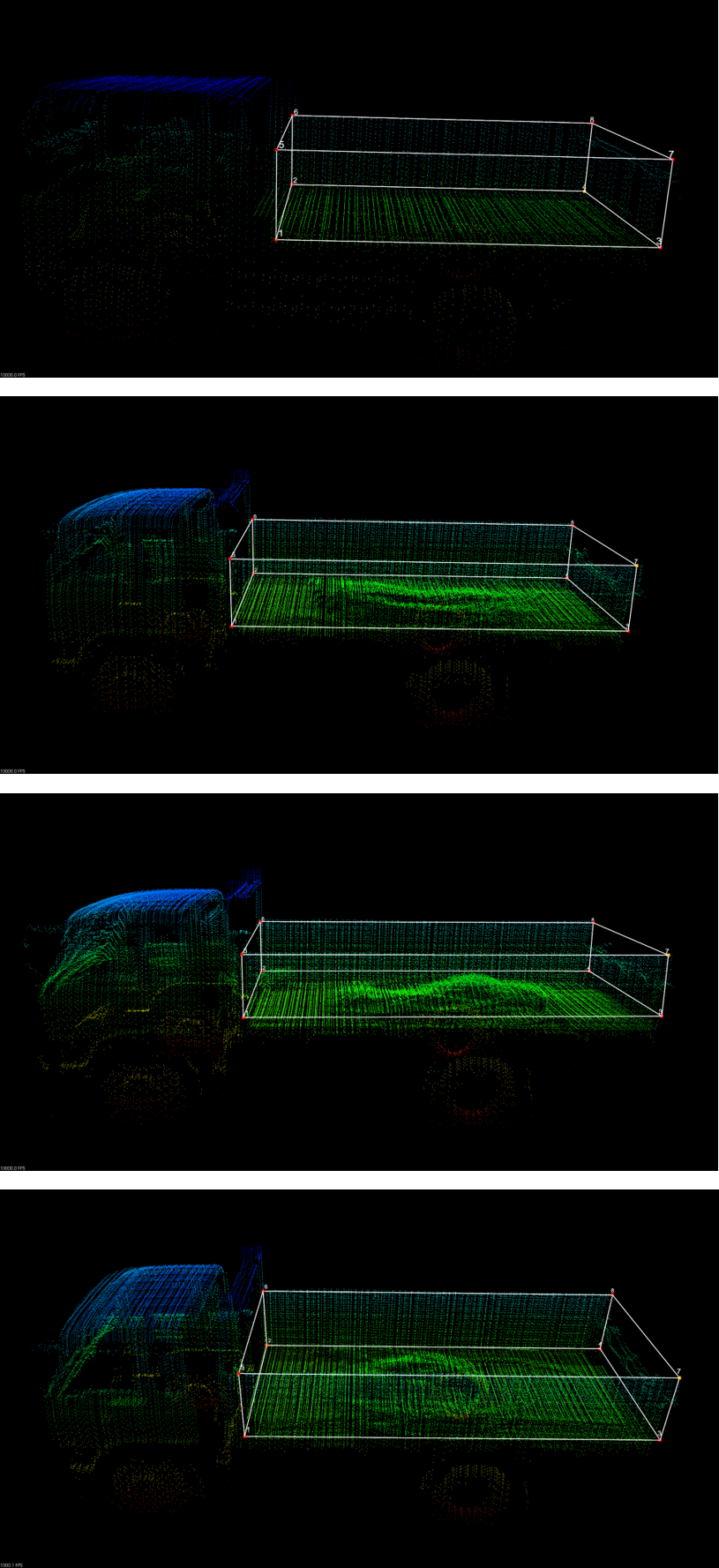}
        }\hspace{-0.7em}
        \subfigure[]{
            \includegraphics[width=1.3in]{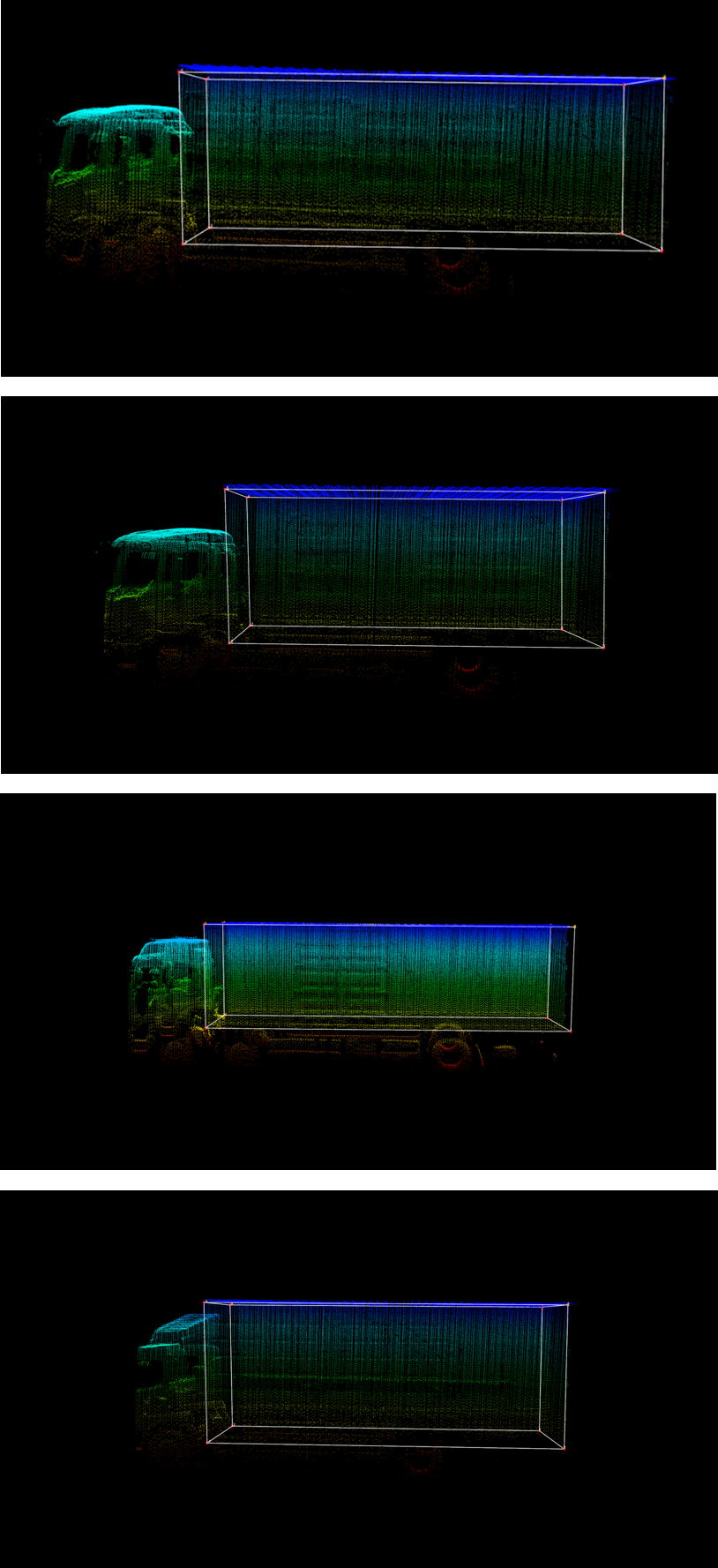}
        }\hspace{-0.7em}
	\caption{Visualization results of our data and public datasets. (a) Our detection results. (b) Large fence truck detection results. (c) Medium fence truck detection results. (d) Small fence truck detection results. (e) Van detection results. The red dot is the key point in locating the compartment, forming the loading space. Our method calculates loading areas that closely align with the actual interior of the compartments, indicating accurate keypoint localization.}
	\label{result}
\end{figure*}

The results of key point localization are shown in Fig. \ref{result}. For visualization purposes, we sequentially connect the eight corner points to form a closed 3-D region. Our collected vehicle point clouds are complete on both the bottom and sides, whereas the ZPVehicles dataset lacks data at the rear of the compartment. Additionally, the interior of the compartments contains foreign objects. As shown in Fig. \ref{result}, the calculated regions closely align with the actual interior of the compartments, indicating that the key point localization is accurate.

To quantitatively analyze the results, we report the detection outcomes for fence trucks collected by LiDAR. As shown in Table \ref{tab:ourdata}, the proposed algorithm demonstrates robustness against the interference of foreign objects within the compartment. However, due to the high-resolution setting of the LiDAR scan, the collected point cloud density is large, resulting in some detection time consumption.
\begin{table}[!t]\color{black}%
	\centering
	\renewcommand{\arraystretch}{1.22}
        \setlength{\tabcolsep}{5pt}
	\caption{Localization results of different datasets and vehicle types}
	\footnotesize
	\begin{tabular}{lccccc}
		\toprule
		\textbf{Type}    &\textbf{\makecell[c]{ADD\\(5\%)}} &\textbf{\makecell[c]{ADD\\(7\%)}} &\textbf{\makecell[c]{ADD\\(10\%)}} &\textbf{\makecell[c]{Average Relative\\ Error(\%)}} &\textbf{\makecell[c]{Runtime\\on CPU(s)}}            \\
		\midrule
		  Truck 	& 0.86  &0.96  &0.98 &-  & 3.66 \\ 
		Van 	  & 0.88  &0.92  &0.94 &-  & 5.08 \\ 
            Truck(Ours data)           & -     &-     &-   `&6.06   & 15.69 \\
		\bottomrule
	\end{tabular}%
	\label{tab:ourdata}%
\end{table}%
\begin{figure}[!t]
	\centering 	
        \subfigure[]{
		\includegraphics[width=1.65in]{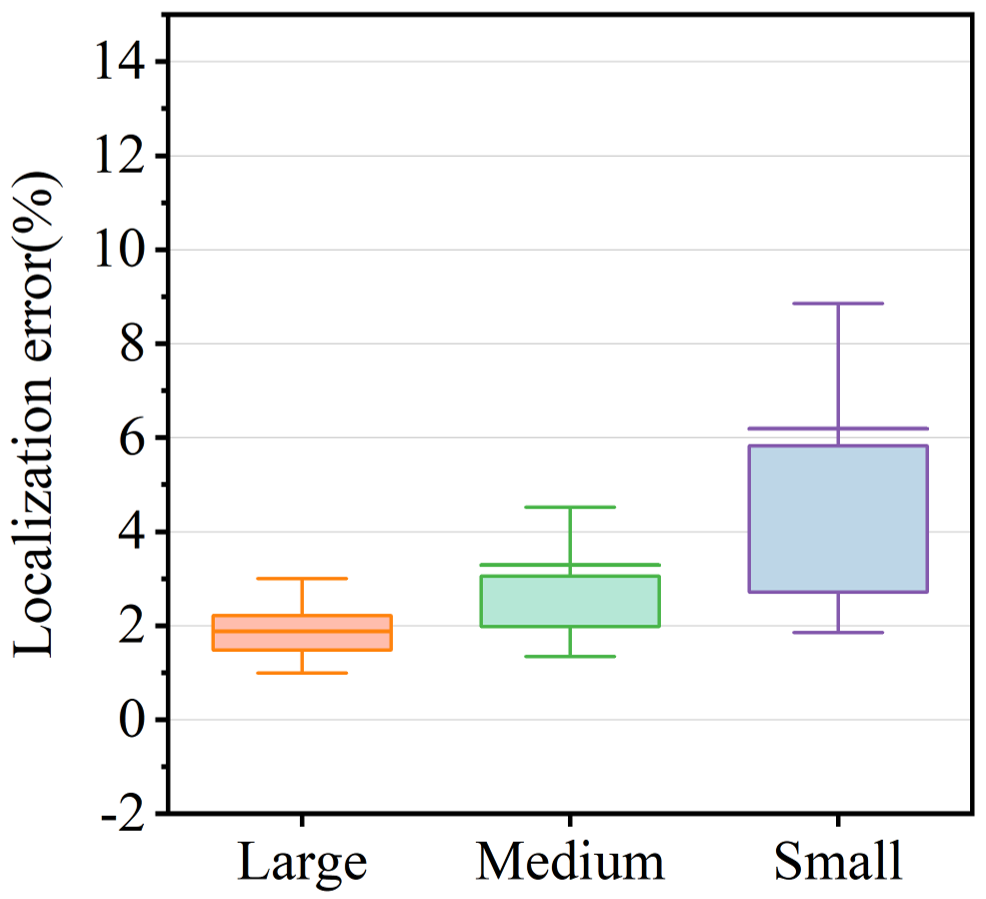}
	}\hspace{-0.7em}
        \subfigure[]{
            \includegraphics[width=1.65in]{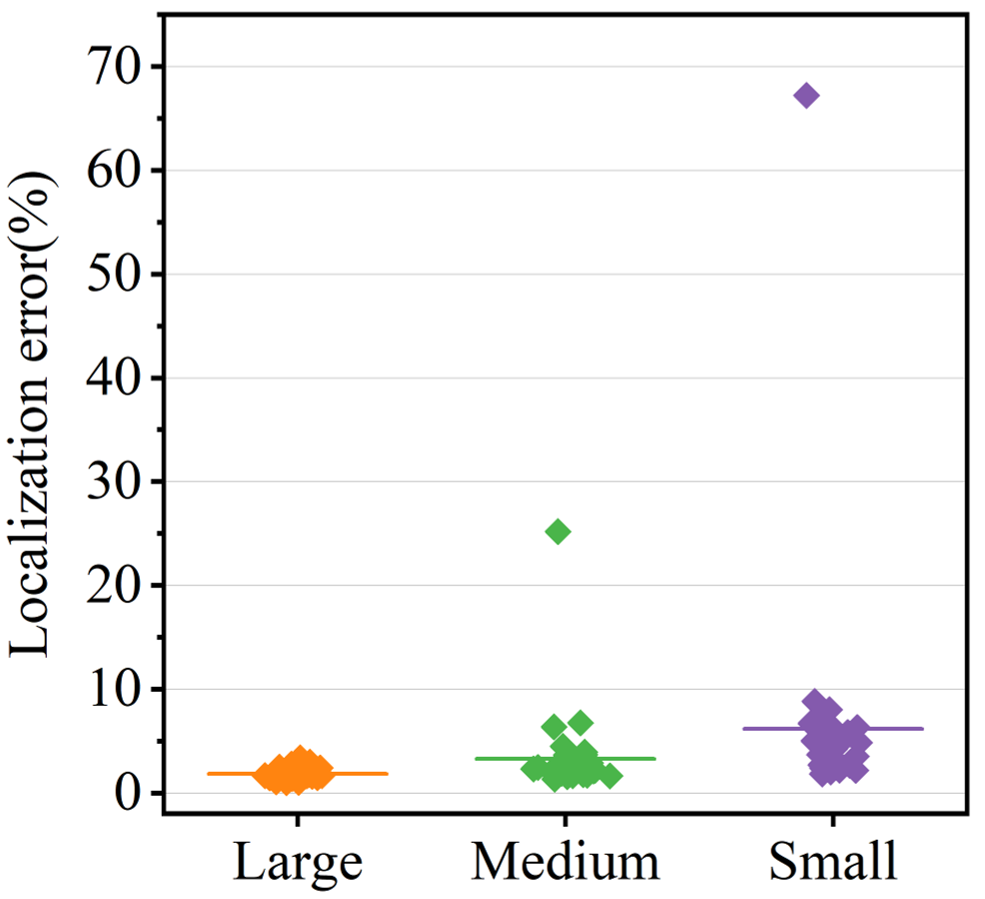}
        }
	\caption{Box and overlap plot of 106 key point localization results set for large, medium, and small vehicles. (a) Boxplot graph. (b) Overlap graph. The line segments within the box represent the mean values. Only 2 vehicles were outliers (exceeding $10\%$), demonstrating that the localization algorithm is robust and accurate.}
	\label{Graph1}
\end{figure}

As illustrated in Fig. \ref{Graph1}, we present the relative error in compartment localization more intuitively using a box plot. Out of 106 vehicles tested, only 2 vehicles had outliers exceeding 10\%. As shown in Table \ref{sixmetrics}, for large vehicles, the mean relative error was 1.88\%, with a standard deviation (SD) of 0.55\%. For medium vehicles, the mean relative error was 3.29\%, with an SD of 3.83\%, demonstrating that the proposed algorithm is robust and accurate. Additionally, on a 6 Core Intel i7 CPU (\textcolor{black}{2.60 GHz}), the average time required to locate key points for 106 vehicles was 3.66 seconds, indicating low computational resource consumption.
\begin{table}[!t]%
	\centering
	\renewcommand{\arraystretch}{1.08}
        \setlength{\tabcolsep}{6pt}
	\caption{Experiment Results of ZPVehicles(Fence Truck) Localization Error(\%)}
	\footnotesize
	\begin{tabular}{cccccc}
		\toprule
		\textbf{Size}  &\textbf{Mean}   &\textbf{SD} &\textbf{Minimum} &\textbf{Median} &\textbf{Maximum}         \\
		\midrule
		Large    & 1.88     & 0.56    & 1.00    & 1.75      & 3.41 \\ 
		  Medium   & 3.29     & 3.83    & 1.34    & 2.38      & 25.22 \\ 
		Small    & 6.19     & 10.77   & 1.85    & 4.21      & 67.24 \\
		\bottomrule
	\end{tabular}%
	\label{sixmetrics}%
\end{table}%

\begin{table*}[!t]\color{black}%
	\centering
	\renewcommand{\arraystretch}{1.18}
    \setlength{\tabcolsep}{16pt}
	\caption{Comparison results with some state-of-the-art methods on ZPVehicles(Fence Truck). Using a two-stage approach, segmentation followed by detection of the bounding box. The positioning error is calculated using the corners of the bounding box. OBB: Oriented Bounding Box}
	\footnotesize
	\begin{tabular}{llccccc}
		\toprule
		\textbf{Vehicles Size}  &\textbf{Methods}   &\textbf{\makecell[c]{ADD\\(5\%)}} &\textbf{\makecell[c]{ADD\\(7\%)}} &\textbf{\makecell[c]{ADD\\(10\%)}} &\textbf{\makecell[c]{Runtime \\on CPU(s)}} &\textbf{\makecell[c]{Runtime \\on GPU and CPU(s)}}         \\
		\midrule
		\multirow{3}*{Large} & PointNet\cite{pointnet} + OBB    & 0.82    & 0.97    & 1.00    & -     & \textbf{0.99} \\ 
		 	        & PointMLP\cite{pointmlp} + OBB   & 0.84     & 0.97    & 1.00    & -   & 0.97 \\ 
		                  & \textbf{Ours}    &\textbf{1.00}     &\textbf{1.00}    & \textbf{1.00}    & 5.53     & - \\
        \multirow{3}{*}{Medium}    & PointNet\cite{pointnet} + OBB    & 0.63    & 0.92    & 0.94    & -     & \textbf{0.63} \\ 
                & PointMLP\cite{pointmlp} + OBB   & 0.65     & 0.94    & 0.94    & -   & 0.63 \\ 
                      & \textbf{Ours}    & \textbf{0.92}     & \textbf{0.97}    & \textbf{0.97}    & 3.62   & - \\
        \multirow{3}{*}{Small}    & PointNet\cite{pointnet} + OBB    & 0.20    & 0.75    & 0.86    & -     & 0.40 \\ 
                & PointMLP\cite{pointmlp} + OBB   & 0.20    & 0.75    & 0.86    & -     & \textbf{0.36} \\ 
                      & \textbf{Ours}    & \textbf{0.62}     & \textbf{0.89}    & \textbf{0.96}    & 1.83     & - \\
		\bottomrule
	\end{tabular}%
	\label{tab:sota}%
\end{table*}%

\begin{figure}[!t]\color{black}
	\centering 	
        \subfigure{
		\includegraphics[width=3.2in]{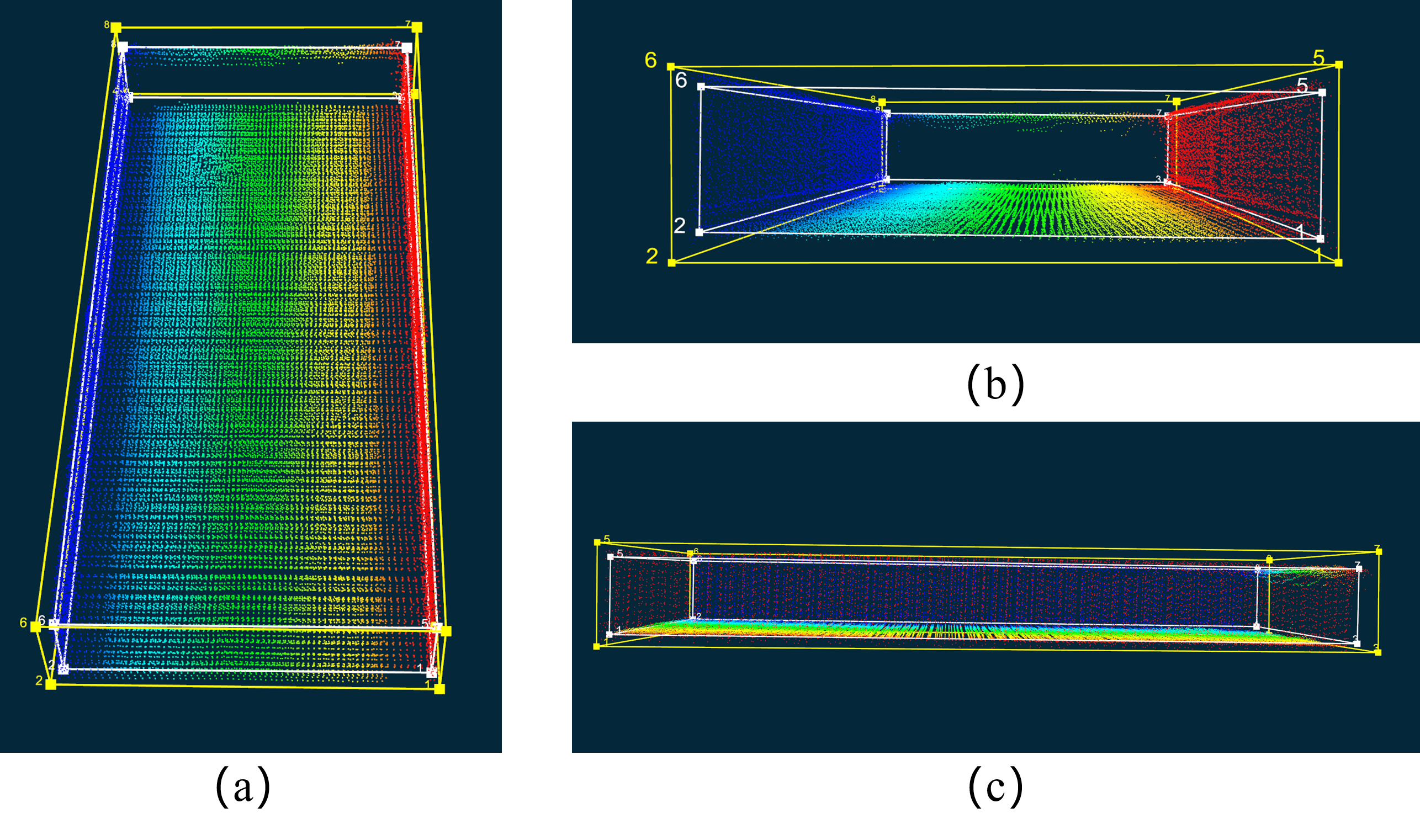}
	}
	\caption{Comparison between 3-D bounding box detection and our method. The yellow box represents the 3-D bounding box detection results, while the white box shows the results of our method. The key points localized by our method are closer to the inner surface of the truck compartment, and the planned area is more suitable for use as the loading and unloading space.}
	\label{sota}
\end{figure}

\textcolor{black}{Previous related research is limited, with the most closely related work to ours focusing on 3-D object detection.  \textcolor{black}{Compared to single-stage methods, segmentation-based models can leverage finer spatial information in the second stage, focusing only on the reduced region of interest point cloud predicted in the first stage, thereby achieving more precise localization and classification\cite{2024survey}. Wang et al. \cite{wang2019research} segment the road scene point cloud into multiple independent clusters and generate bounding boxes for the segmented point cloud clusters to locate the 3D spatial positions of vehicles. The method \cite{subobb, chen2019novel} also adopts point cloud segmentation and oriented bounding boxes (OBB) to estimate the spatial location information of objects. To further demonstrate the superiority of our method, we compared it with 3-D object detection methods based on point cloud segmentation.} The results are shown in Table \ref{tab:sota}, which compares segmentation accuracy and detection time. First, the two-stage method runs on a GPU, providing faster detection speed but lacking localization accuracy. This is due to the irregular shape and thickness of the truck compartment, causing the detected 3-D bounding box corners to deviate from the compartment. In contrast, the key points localized by our method are closer to the compartment, thanks to the use of geometric features and prior knowledge of the vehicle, as shown in Fig. \ref{sota}. Moreover, relying on a GPU-based method significantly increases device costs, making it unsuitable for large-scale deployment.}

\begin{table}[!t]%
	\centering
	\renewcommand{\arraystretch}{1.08}
        \setlength{\tabcolsep}{7pt}
	\caption{Localization Results of ZPVehicles(Fence Truck)}
	\footnotesize
	\begin{tabular}{cccccc}
		\toprule
		\textbf{Methods}  &\textbf{Size}   &\textbf{\makecell[c]{ADD\\(5\%)}} &\textbf{\makecell[c]{ADD\\(7\%)}} &\textbf{\makecell[c]{ADD\\(10\%)}} &\textbf{\makecell[c]{Runtime\\on CPU(s)}}         \\
		\midrule
		\multirow{4}*{No Optimize} & Large    & 0.33    & 0.48    & 0.56    & 5.49 \\ 
		 	        & Medium   & 0.34     & 0.44    & 0.52    & 3.58 \\ 
		                  & Small    & 0.03     & 0.10    & 0.13    & 1.80 \\
                            & All   &0.23   &0.34   &0.40   &\textbf{3.62}\\
            \multirow{4}{*}{Optimize}    & Large    & 1.00    & 1.00    & 1.00    & 5.53 \\ 
		 	        & Medium   & 0.92     & 0.97    & 0.97    & 3.62 \\ 
		                  & Small    & 0.62     & 0.89    & 0.96    & 1.83 \\
                        & All   &\textbf{0.86}   &\textbf{0.96}   &\textbf{0.98}   &3.66 \\
		\bottomrule
	\end{tabular}%
	\label{tab:opt}%
\end{table}%

\begin{figure}[!t]\color{black}
	\centering 	
        \subfigure[]{
		\includegraphics[width=1.6in]{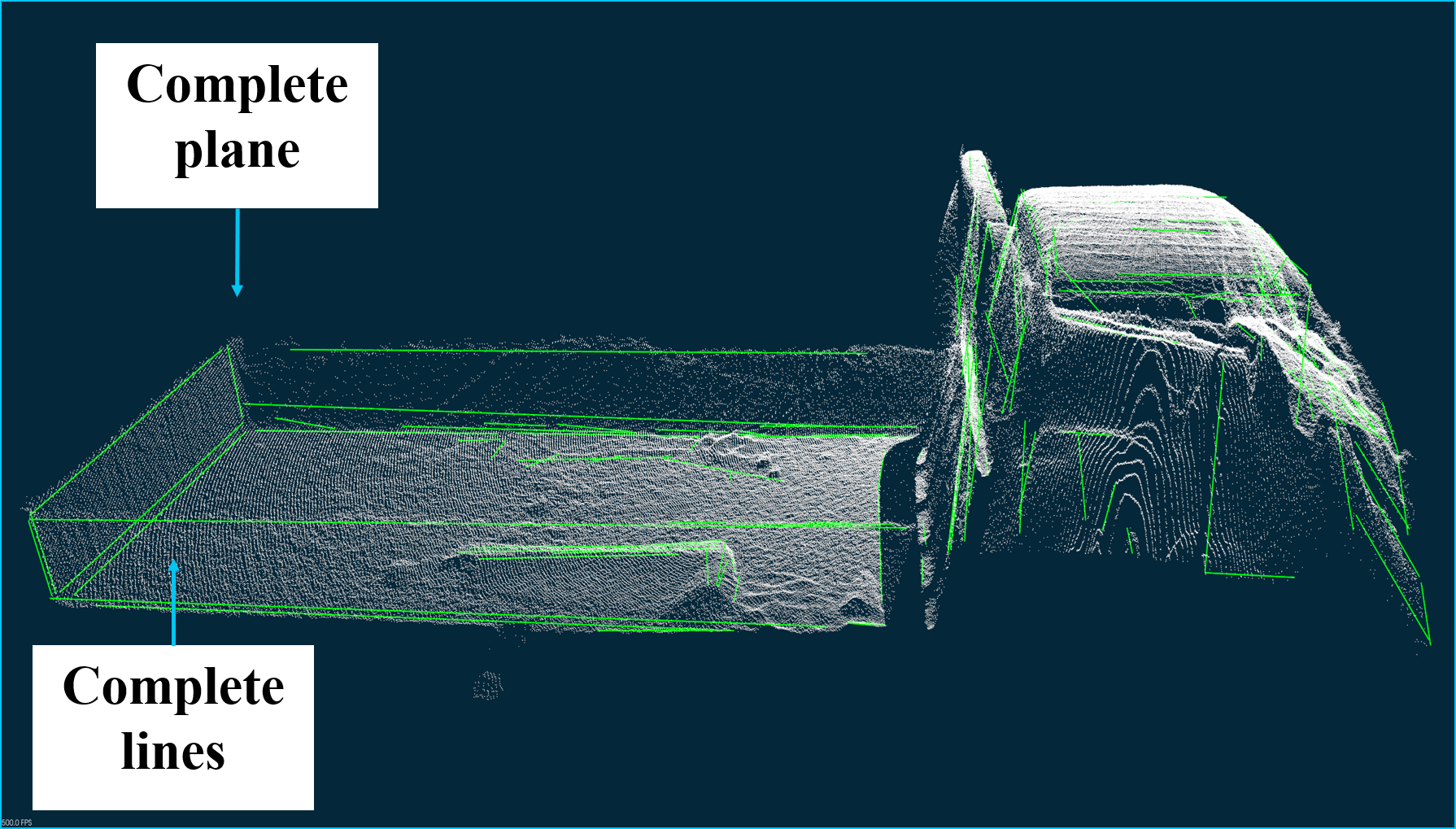}
	}\hspace{-0.7em}
        \subfigure[]{
            \includegraphics[width=1.6in]{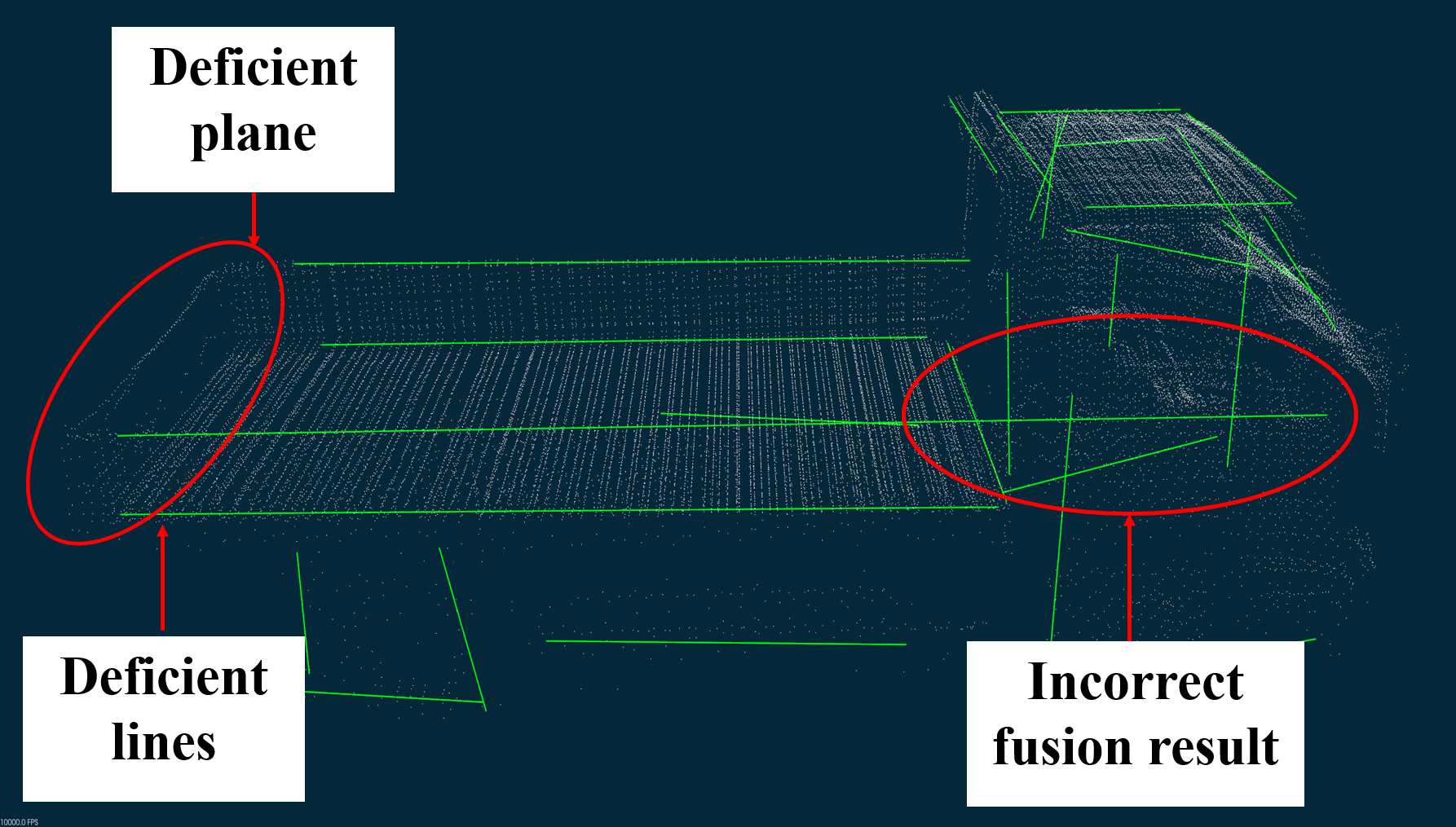}
        }\hspace{-0.7em}
        \subfigure[]{
            \includegraphics[width=1.6in]{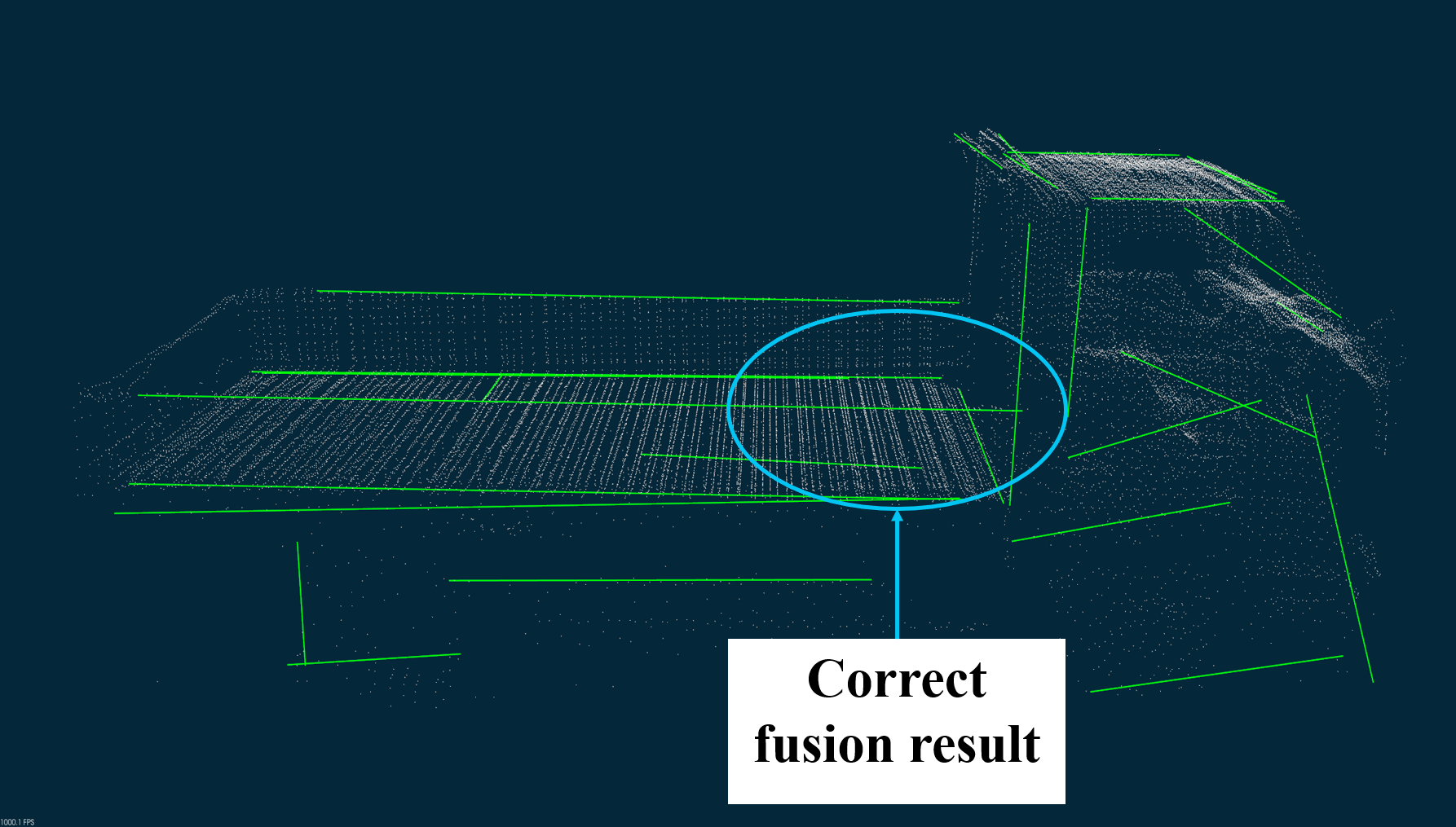}
        }\hspace{-0.7em}
        \subfigure[]{
            \includegraphics[width=1.6in]{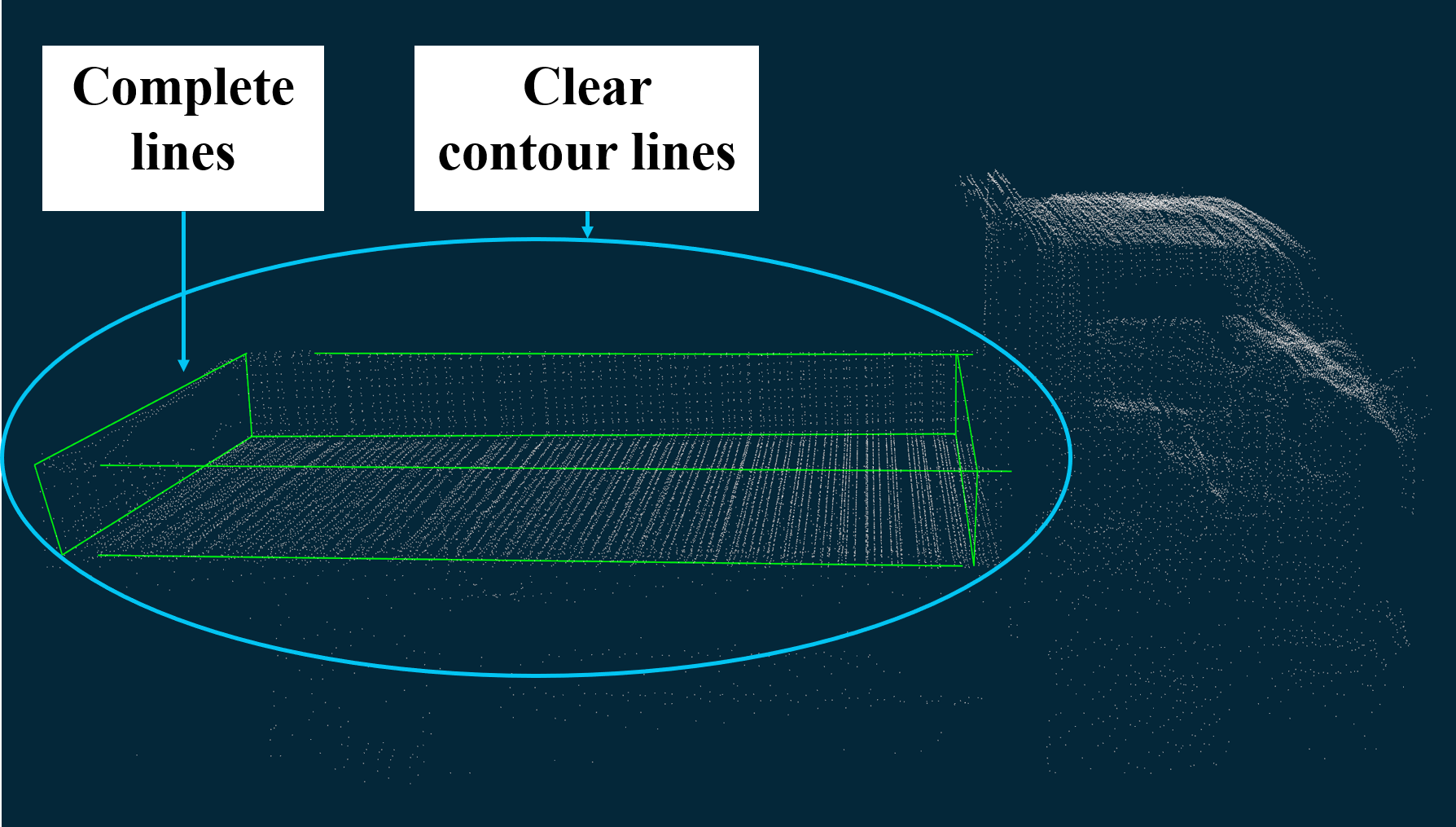}
        }
	\caption{Contour line optimization. (a) Our Datasets Detection Results. (b) 3-D segments detected by the baseline method. (c) Optimized contour lines. (d) Contour Line Optimization Final Result.}
	\label{opt}
\end{figure}

\subsection{\textcolor{black}{Ablation Study}}
\subsubsection{Optimization of Truck Contour}
Through the vehicle point cloud measured by this device, the information on each side of the truck is complete, as shown in Fig. \ref{opt}(a), and the detected three-dimensional line segment can better reflect the geometric contour of the truck compartment. The detection result of the method \cite{ref20} is shown in Fig. \ref{opt}(b). Due to the vehicle point cloud provided by ZPVehicles lacking information behind the vehicle compartment, the detected line segments are incomplete. Using the method in section \uppercase\expandafter{\romannumeral3}, the detected line segments are fused in their respective planes to obtain the results shown in Fig. \ref{opt}(c), and then the contour optimization and completion strategy are implemented. As shown in Fig. \ref{opt}(d), our method can obtain clearer and more complete features of the truck edge line and can provide accurate contour information for the key point location in the next stage.

To validate the effectiveness of the contour optimization strategy, we conducted ablation experiments on the ZPVehicles dataset. As shown in Table \ref{tab:opt}, after applying the optimization strategy, the detection error for large vehicles remained below the 5\% threshold, and the localization accuracy for medium and small vehicles significantly improved without impacting the algorithm detection speed. 

\subsubsection{\textcolor{black}{Hyperparameter}}
\begin{figure}[!t]\color{black}
	\centering 	
        \subfigure[]{
		\includegraphics[width=1.65in]{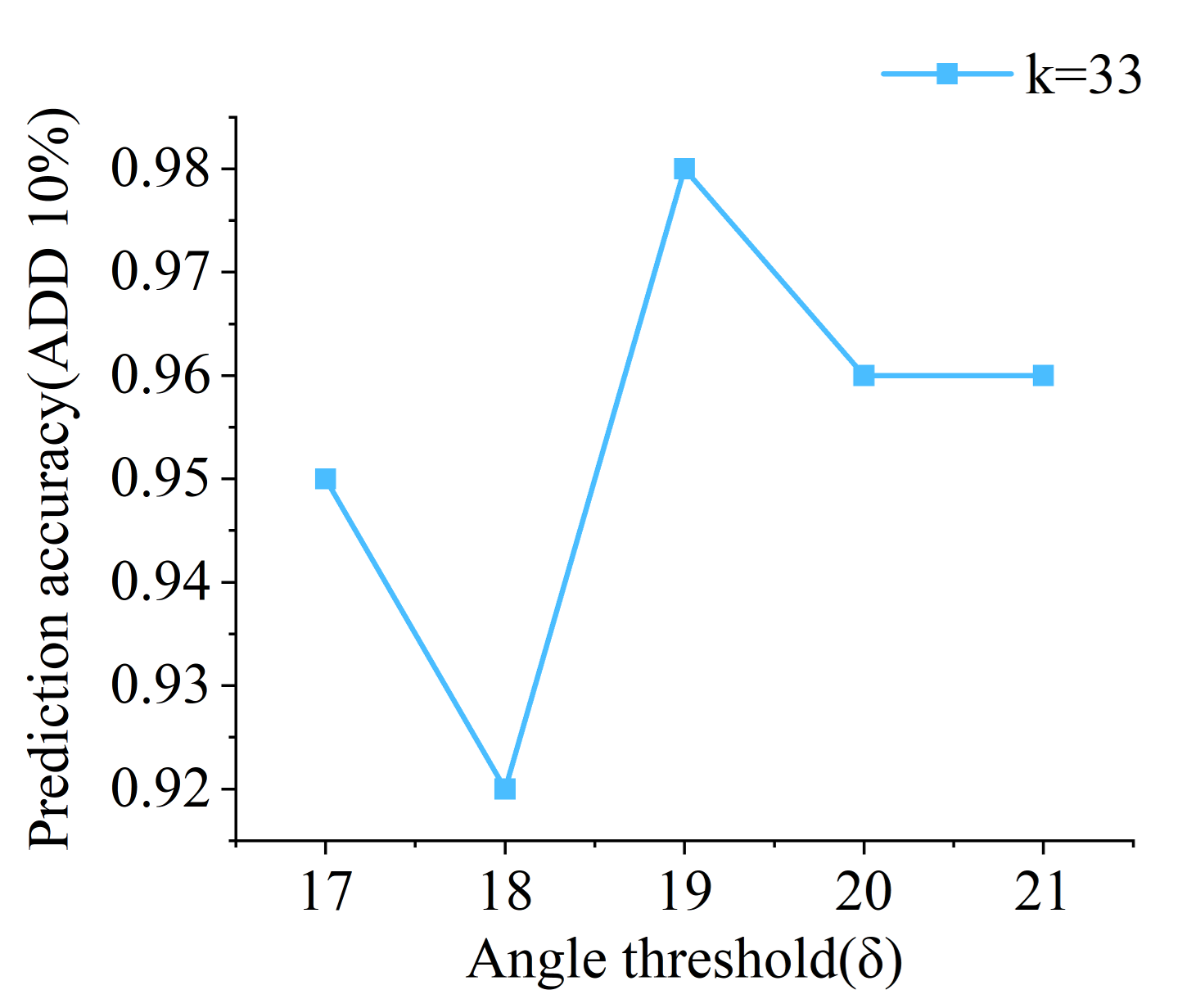}
	}\hspace{-0.7em}
        \subfigure[]{
            \includegraphics[width=1.65in]{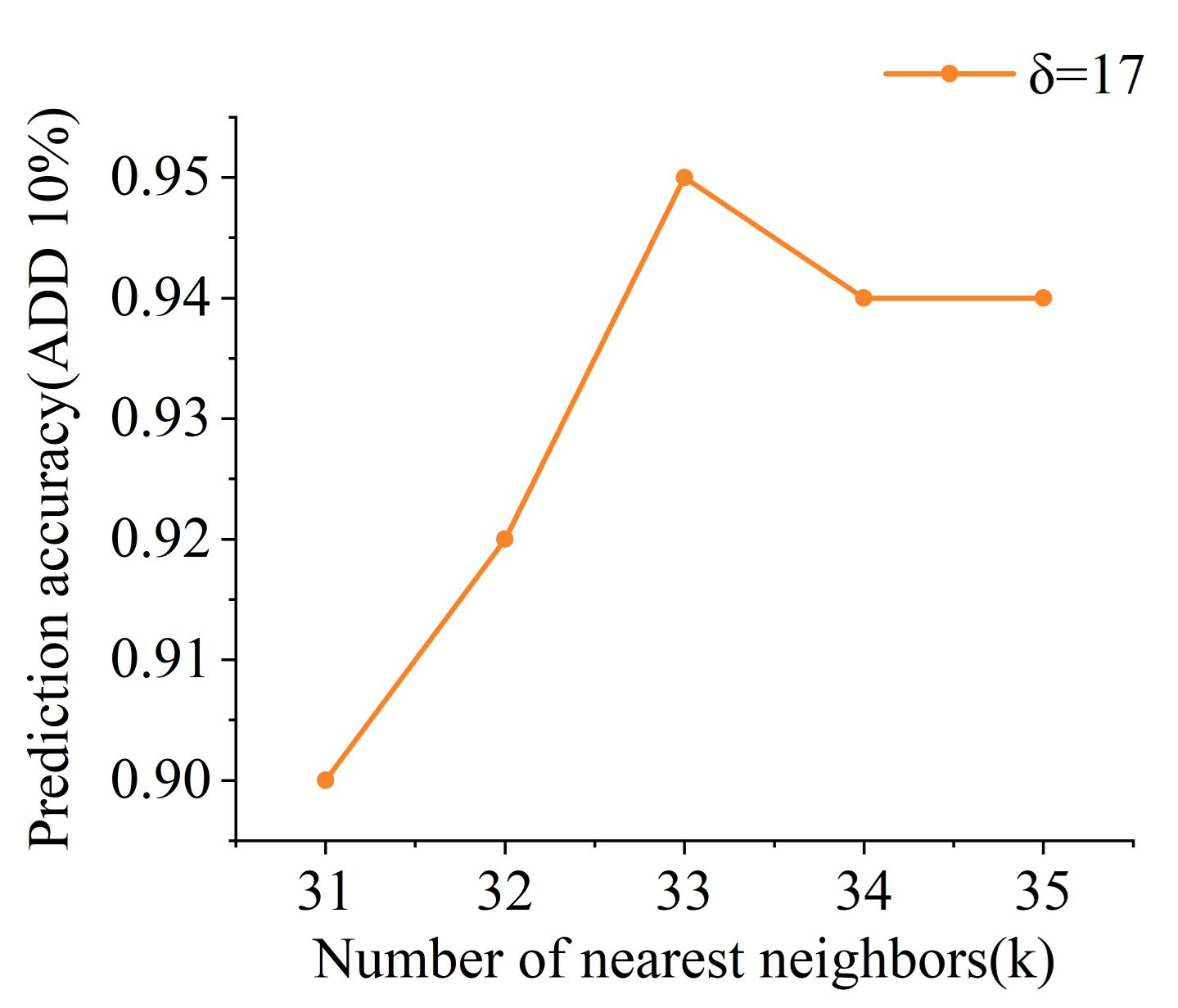}
        }
	\caption{\textcolor{black}{Hyperparameter experiment. (a) Effect of hyperparameter $k$ on experimental results. (b) Effect of hyperparameter $\delta$ on experimental results.}}
	\label{hyper}
\end{figure}

\begin{figure}[!t]\color{black}
	\centering 	
        \subfigure[]{
		\includegraphics[width=1.65in]{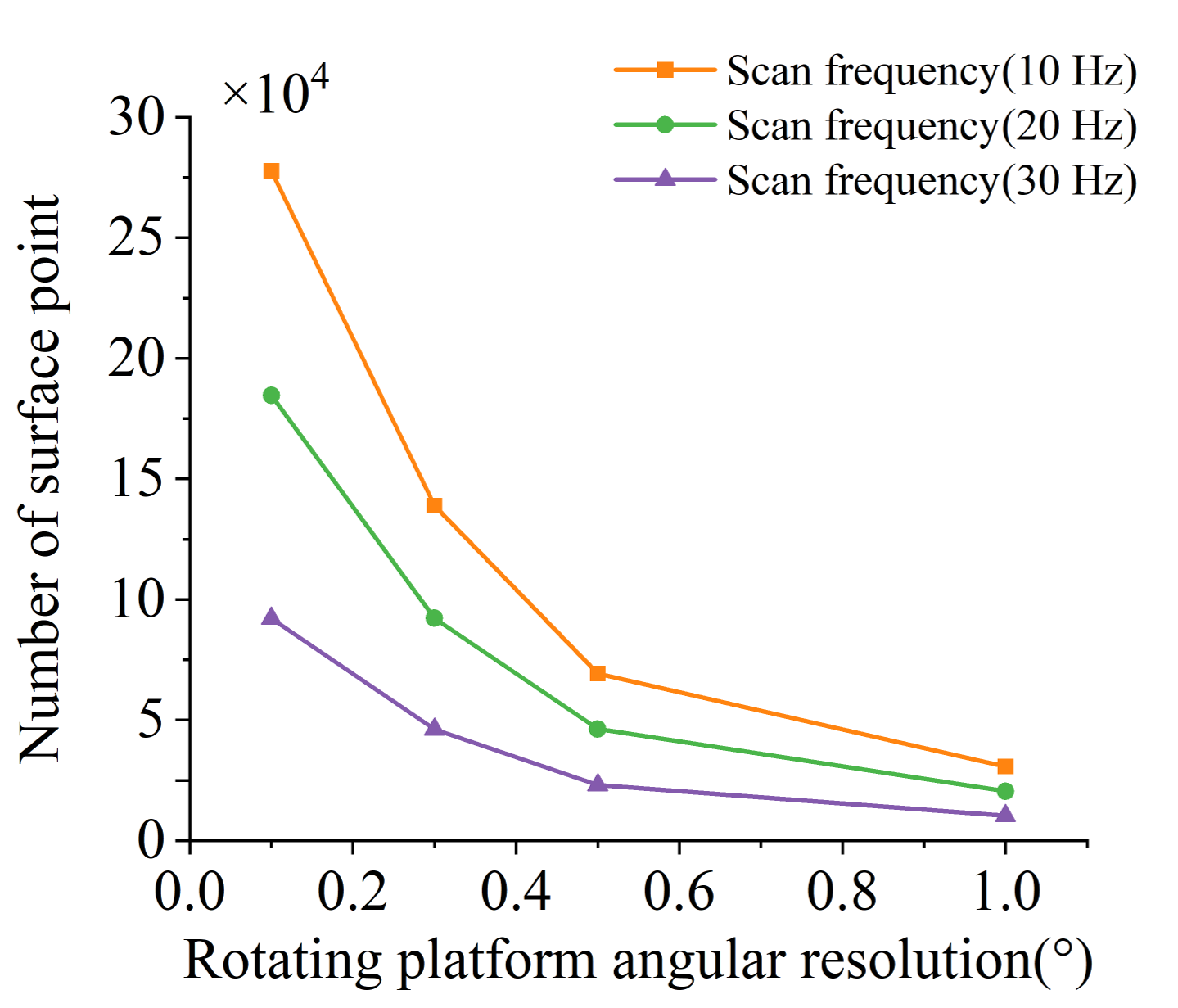}
	}\hspace{-0.7em}
        \subfigure[]{
            \includegraphics[width=1.65in]{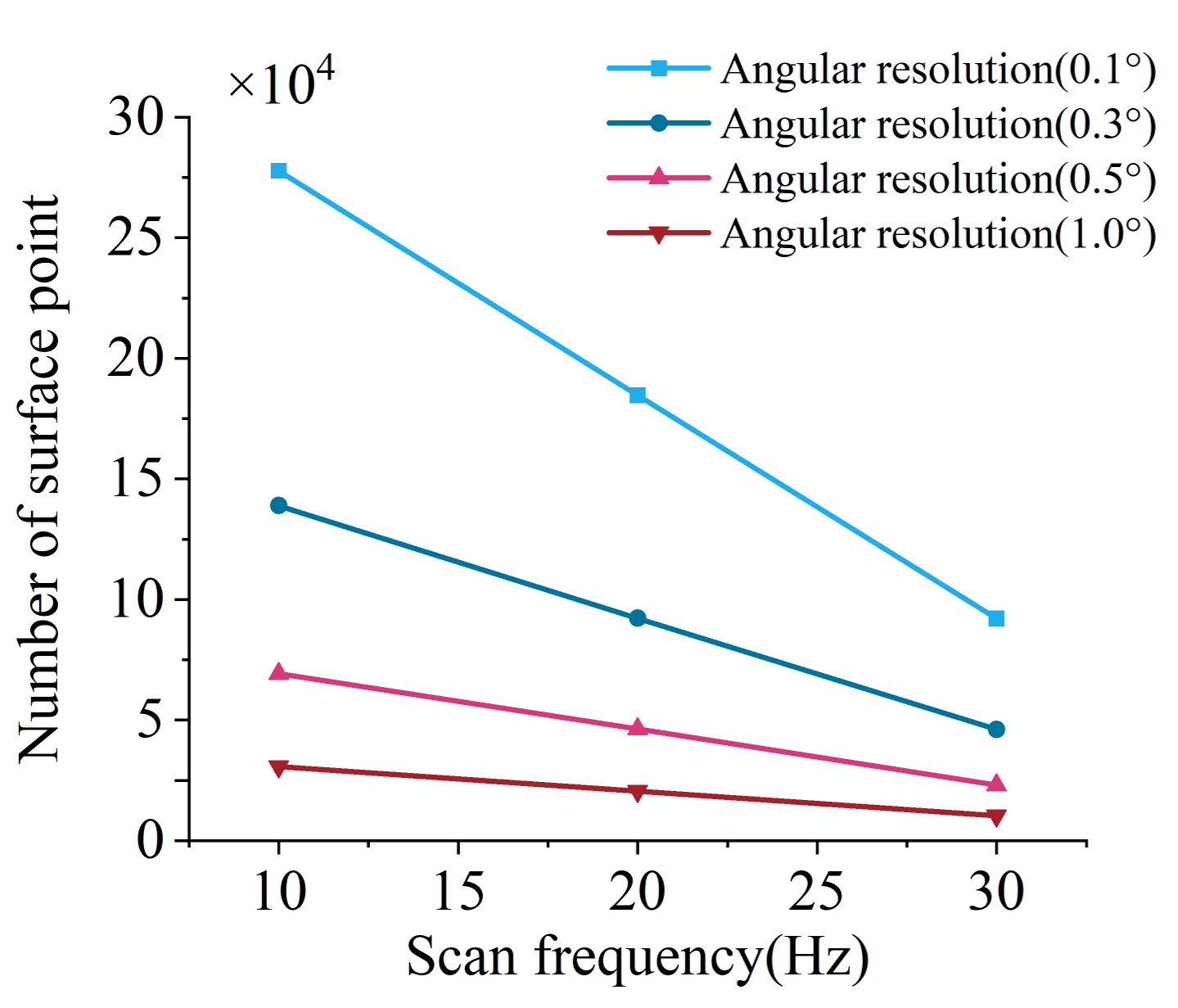}
        }
	\caption{\textcolor{black}{System parameter experiment. (a) Effect of rotating platform resolution on data acquisition. (b) Effect of 2-D LiDAR scan frequency on data acquisition. Fine angular resolution and low scanning frequency will capture more points on the vehicle's surface but will increase the scanning and processing time.}}
	\label{lidar_cfg}
\end{figure}
\textcolor{black}{\textcolor{black}{The proposed key point localization algorithm for the truck compartment depends on the point cloud segmentation results of the compartment plane. The method \cite{ref20} extracts small, compact regions first and then merges these regions to obtain the final result, with its performance heavily influenced by the parameters $k$ and $\delta$. To explore the impact of point cloud segmentation parameters on localization accuracy, we designed and conducted an ablation experiment.} As shown in Fig. \ref{hyper}, with the hyperparameter $\delta$ fixed, the prediction accuracy of the algorithm gradually increases as $k$ increases, reaching a local optimum at $k=33$. Meanwhile, increasing $\delta$ leads to the model achieving global optimal accuracy at $\delta=19$. More detailed data can be found in Tables \ref{tab:hyper_k} and \ref{tab:hyper_d}. \textcolor{black}{The method \cite{pcl} provides an explanation for this phenomenon. In region-growing segmentation, a larger $k$ value expands the neighborhood range, allowing more points to participate in the growth of the region. However, an excessively large $k$ value may cause different regions to be incorrectly merged. A larger normal vector angle threshold $\delta$ relaxes the segmentation conditions, which helps the algorithm better handle point cloud regions with significant normal variations. However, if the threshold is set too large, irrelevant points may be incorrectly assigned to the same region. Therefore, excessively large $k$ and $\delta$ values can have a severe impact on the detection results.}}

\subsubsection{\textcolor{black}{System configuration}}
\textcolor{black}{\textcolor{black}{Equation (1) is determined by the angular resolution of the rotating platform and the scanning frequency of the 2-D LiDAR. The former directly affects the spatial distribution of each point during the scanning process, while the latter determines the number of points collected per unit of time. These factors influence the point cloud density and the level of detail. To explore the impact of LiDAR scanning parameters on data collection, we designed and conducted ablation experiments.} As shown in Fig. \ref{lidar_cfg}, a smaller angular resolution and a lower scan frequency result in a higher number of points on the vehicle surface, providing clearer details of the vehicle's local features. \textcolor{black}{However, collecting points from more locations in the scanning space will introduce additional computational overhead in Equation (1), which significantly increases the time cost of data collection and processing. Therefore, to balance point cloud quality, scanning efficiency, and time cost, we chose to set the angular resolution of the rotating platform to $0.3^\circ$ and the scanning frequency of the 2-D LiDAR to $25$ Hz. This configuration reduces the scanning time while also achieving good point cloud quality.}}

\begin{figure}[!t]\color{black}
	\centering 	
        \subfigure{
		\includegraphics[width=2.9in]{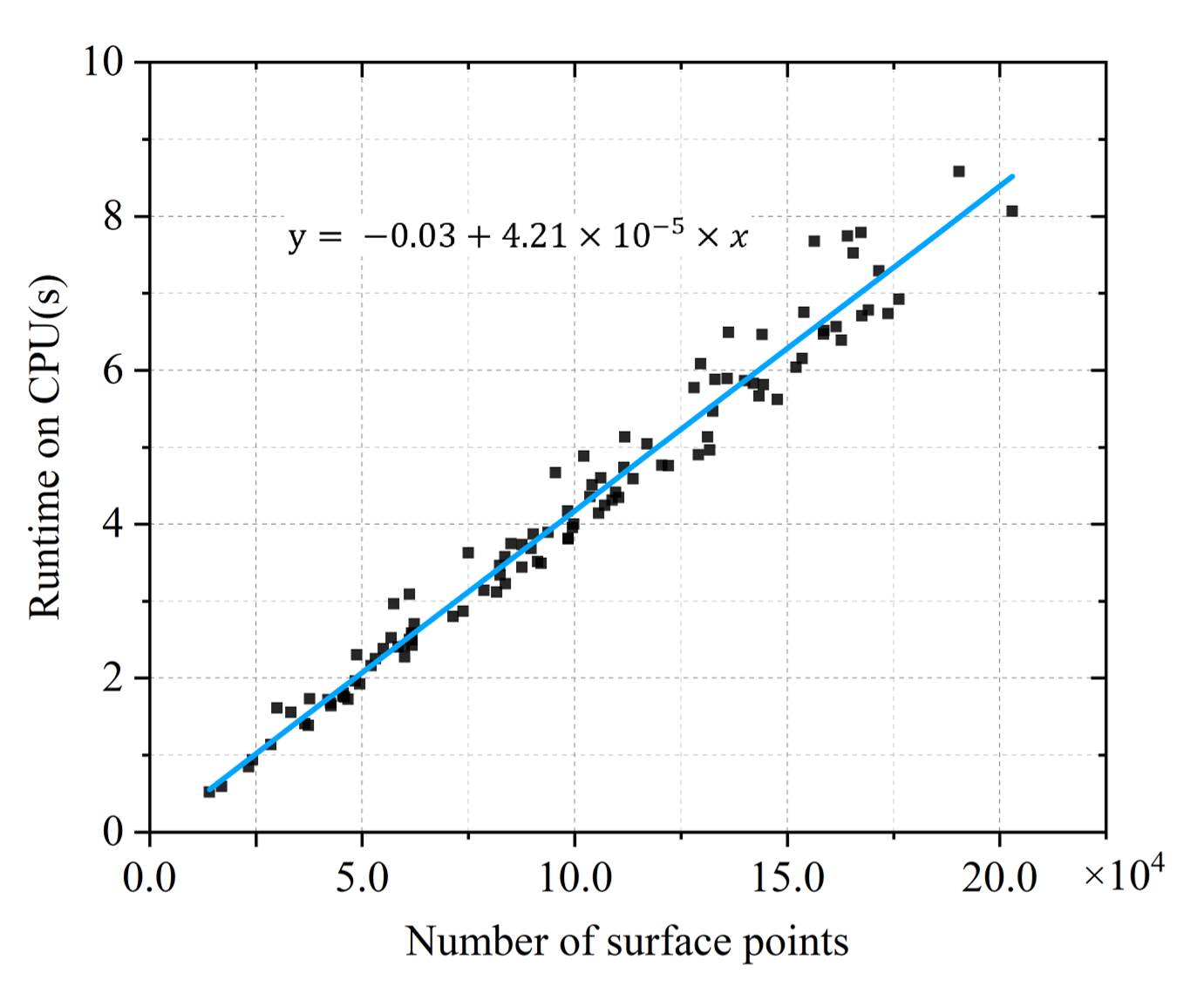}
	}
	\caption{The relationship between the number of vehicle point clouds and detection time on the CPU. Black dots represent individual vehicles, while the blue line is a fitted curve based on the distribution trend of the points, indicating a near-linear relationship between the number of vehicle point clouds and detection time. The average detection time for large vehicles is 5.53 seconds, which meets the requirements of practical applications.}
	\label{CPU}
\end{figure}

\section{\textcolor{black}{Discussion and Limitation}}
\textcolor{black}{This paper introduces a LiDAR-based localization system that can automatically locate the position of fence truck compartments of different sizes and demonstrate reliability in cluttered scenes. In practical applications, we used LiDAR to collect point cloud data from commonly used fenced trucks and conducted localization tests. The average localization error was 6.06\%. On a public dataset, the localization error was 5\%, with an accuracy of 86\%; when the localization error was 7\%, the accuracy increased to 96\%. Data processing was performed on a 6-core Intel i7 CPU (2.60 GHz) without a GPU. The average detection time for large vehicles is 5.53 seconds, which meets the requirements of practical applications. Figures \ref{lidar_cfg} and \ref{CPU} demonstrate that the detection time increases linearly as the number of vehicle point clouds grows. Additionally, our developed device can transmit data to different terminals via Ethernet, facilitating integration with third-party devices.}

\begin{table}[!t]\color{black}
	\centering
	\renewcommand{\arraystretch}{1.08}
        \setlength{\tabcolsep}{14pt}
	\caption{Ablation Study of Hyperparameter $k$}
	\footnotesize
	\begin{tabular}{ccccc}
		\toprule
		\bm{$k$} &\textbf{\makecell[c]{ADD\\(5\%)}} &\textbf{\makecell[c]{ADD\\(7\%)}} &\textbf{\makecell[c]{ADD\\(10\%)}} &\textbf{\makecell[c]{Runtime\\on CPU(s)}}         \\
		\midrule
		31  & 0.81  & 0.87  & 0.90  & \textbf{3.92} \\ 
		  32  & 0.82  & 0.89  & 0.92  & 4.05 \\ 
		\textbf{33}  & 0.87  & 0.91  & \textbf{0.95}  & 4.15 \\
            34  & 0.85  & 0.91  & 0.94  & 4.17 \\
            35  & 0.85  & 0.92  & 0.94  & 4.23 \\
		\bottomrule
	\end{tabular}%
	\label{tab:hyper_k}%
\end{table}%

\begin{table}[!t]\color{black}
	\centering
	\renewcommand{\arraystretch}{1.08}
        \setlength{\tabcolsep}{14pt}
	\caption{Ablation Study of Hyperparameter $\delta$}
	\footnotesize
	\begin{tabular}{ccccc}
		\toprule
		\bm{$\delta$} &\textbf{\makecell[c]{ADD\\(5\%)}} &\textbf{\makecell[c]{ADD\\(7\%)}} &\textbf{\makecell[c]{ADD\\(10\%)}} &\textbf{\makecell[c]{Runtime\\on CPU(s)}}         \\
		\midrule
		17  & 0.87  & 0.91  & 0.95  & 4.15 \\ 
		  18  & 0.83  & 0.91  & 0.92  & 4.46 \\ 
		\textbf{19}  & 0.86  & 0.96  & \textbf{0.98}  & 4.04 \\
            20  & 0.85  & 0.93  & 0.96  & 4.25\\
            21  & 0.86  & 0.94  & 0.96  &\textbf{4.00}\\
		\bottomrule
	\end{tabular}%
	\label{tab:hyper_d}%
\end{table}%

\begin{figure}[!t]\color{black}
	\centering 	
        \subfigure[]{
		\includegraphics[width=3.0in]{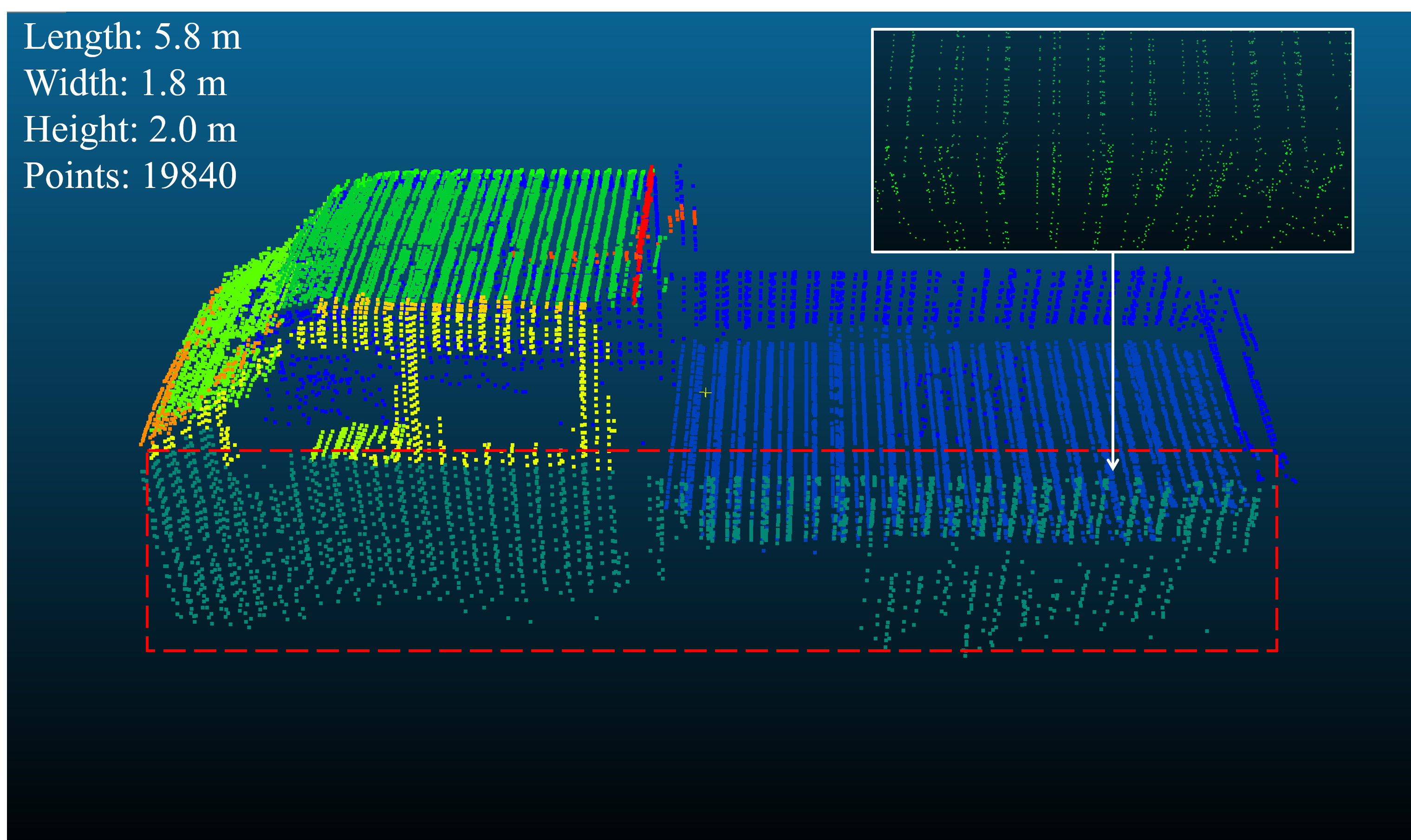}
	}\hspace{-0.7em}
        \subfigure[]{
            \includegraphics[width=3.0in]{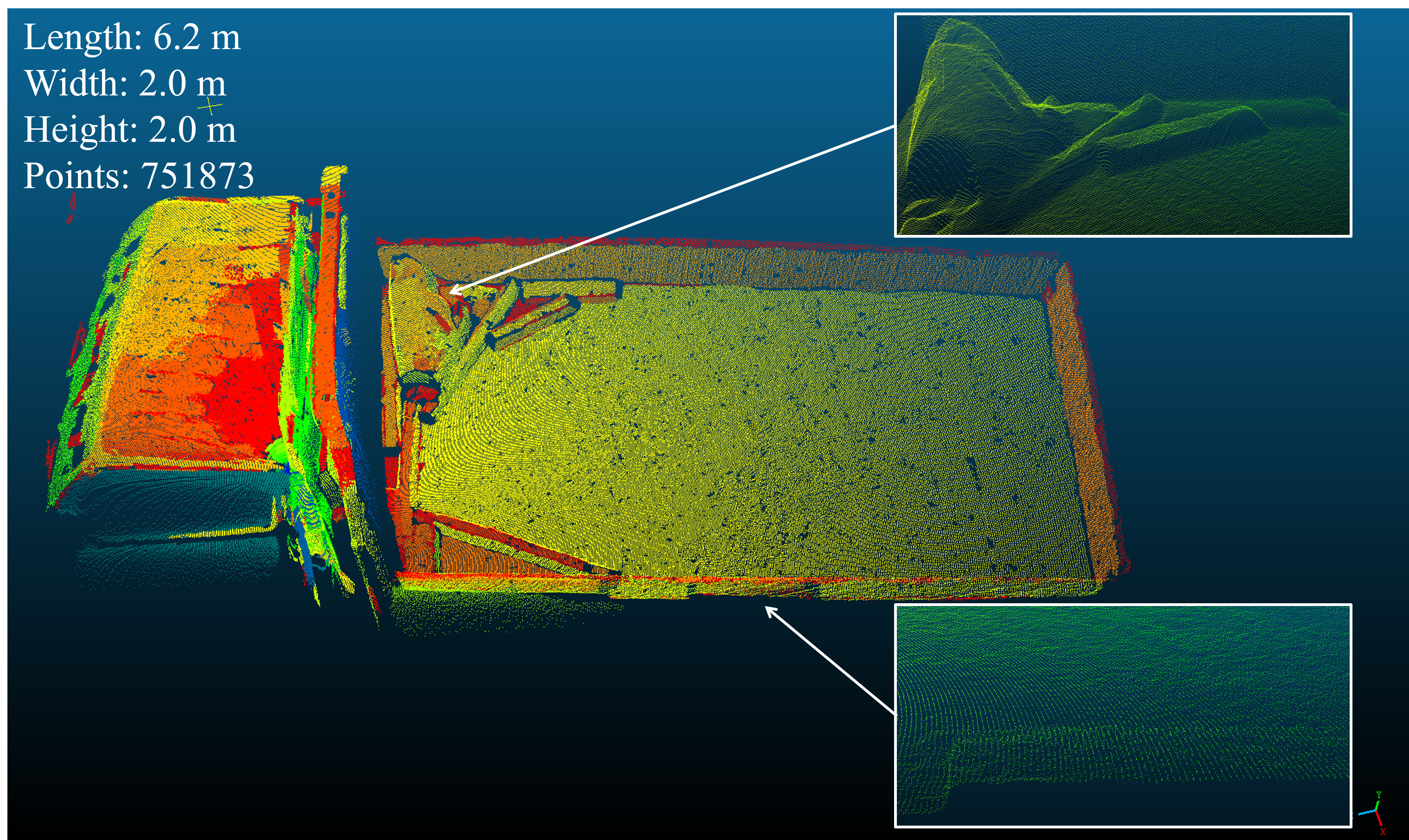}
        }
	\caption{\textcolor{black}{Abnormal situation analysis. (a) Segmentation results of public dataset. (b) Segmentation results of our dataset. Different colors represent different segmentation categories. The point cloud from the public dataset is unevenly distributed and severely sparse, leading to over-segmentation. In comparison, our dataset has higher point cloud density and clearer local details, resulting in better segmentation outcomes.}}
	\label{badcase}
\end{figure}

The contour optimization strategy we propose is robust against missing point clouds at the rear of the vehicle and foreign objects inside the truck compartments. \textcolor{black}{For large vehicles, the prediction results of this method are nearly entirely accurate, while for medium and small vehicles, the recognition results of two vehicles are considered outliers (exceeding 10\%). To further analyze the shortcomings of this method in positioning small vehicles, we visualized the detection process.} Region-growing segmentation treats the side of the truck compartment and the front of the vehicle as a single plane, as highlighted by the red dashed box in Fig. \ref{badcase}(a). This causes the planned space to extend beyond the truck compartment, leading to significant measurement errors. Observing the local details of the point cloud of the carriage, the point cloud distribution is uneven and the sparsity is serious, which makes the regional growth segmentation unable to determine the segmentation edge, resulting in excessive segmentation. \textcolor{black}{By comparison, the data collected by our device is clearer in terms of point cloud density and geometric contours. The same method achieved better detection results, as shown in Fig. \ref{badcase}(b) and Table \ref{tab:ourdata}.} \textcolor{black}{To address the challenges posed by uneven point cloud distribution and sparsity, future work will focus on improvements in both point cloud segmentation algorithms and data processing. \textcolor{black}{In terms of algorithms, inspired by the latest advancements \cite{ptv3}, we recognize that ordering point cloud data based on specific patterns and converting it into structured sequences can achieve a balance between the accuracy and efficiency of 3-D point cloud segmentation. In terms of data processing, we will further investigate methods for recovering dense and evenly distributed point clouds from sparse and noisy data, aiming to reduce errors caused by sensor discrepancies. The Transform framework, with its powerful global preservation and precise local detail capture, has been widely applied to point cloud completion tasks\cite{adapointr}\cite{pointcformer} and will be a key focus of our next steps.}}

\textcolor{black}{However, this study has certain limitations. Our system is best suited for precise static measurements rather than real-time applications. The relative position between the LiDAR and the vehicle must remain stationary. In this way, it constructs an accurate 3-D map by combining point clouds scanned from different positions, rather than using SLAM technology. Additionally, environmental vibration noise and the size of the device limit its deployment in certain specific spaces or preferred lightweight equipment \cite{cui}. Future work will focus on addressing these limitations by enhancing the portability of the system and exploring real-time application techniques to handle more dynamic scenarios. }

\section{Conclusion}
This paper studies the localization of fence truck compartments in automatic loading systems and explores a wide field-of-view 3-D LiDAR compartment automatic localization system. Specifically, the system comprises both hardware and software modules: a rotating LiDAR generates high-density vehicle point clouds over a wide field of view, adapting to uneven lighting environments. The software module designs a method that simultaneously establishes a global coordinate system and segments the vehicle point clouds. The designed vehicle key point localization algorithm directly processes the vehicle point clouds and outputs accurate compartment key points, providing usable loading and unloading spaces. Extensive experiments on large, medium, and small fence trucks demonstrate that the system offers high localization accuracy and lower computational resource consumption.

Future work will focus on several directions: we will explore more types of prior information and study how to effectively integrate them into the key point localization algorithm to accommodate various vehicle types. Additionally, we will use smaller and better-performing sensors to enhance the portability and usability of the system \textcolor{black}{and exploring real-time application techniques to handle more dynamic scenarios.} 

\begin{thebibliography}{99}  
\bibliographystyle{IEEEtran}
\bibitem{ref1}
D.~Shen, J.~Hu, T.~Zhai, T.~Wang, and Z.~Zhang, ``Parallel loading and unloading: smart technology towards intelligent logistics,'' in \emph{IEEE International Conference on Systems, Man and Cybernetics}, 2019, pp. 847--851.

\bibitem{ref2}
L.~Han, C.~Ruijuan, and M.~Enrong, ``Design and simulation of a handling robot for bagged agricultural materials,'' \emph{IFAC-PapersOnLine}, vol.~49, no.~16, pp. 171--176, 2016.

\bibitem{ref3}
F.~Ma, B.~Zhang, P.~Ji, Y.~Sun, J.~Li, X.~Li, W.~Yan, F.~Wang, and G.~Mu, ``Research on positioning and dimension measurement of truck compartment based on semantic segmentation in automatic loading system,'' in \emph{Proceedings of the 2023 6th International Conference on Signal Processing and Machine Learning}, 2023, pp. 89--96.

\bibitem{ref4}
J.~Mrovlje and D.~Vran{\v{c}}i{\'c}, ``Automatic detection of the truck position using stereoscopy,'' in \emph{IEEE International Conference on Industrial Technology}, 2012, pp. 755--759.

\bibitem{ref5}
W.~Zhang, L.~Gong, S.~Chen, W.~Wang, Z.~Miao, and C.~Liu, ``Autonomous identification and positioning of trucks during collaborative forage harvesting,'' \emph{Sensors}, vol.~21, no.~4, p. 1166, 2021.

\bibitem{ref6}
W.~Zou, D.~Shen, P.~Cao, C.~Lin, and J.~Zhu, ``Fast positioning method of truck compartment based on plane segmentation,'' \emph{IEEE Journal of Radio Frequency Identification}, vol.~6, pp. 774--778, 2022.

\bibitem{ref7}
M.~Wei, J.-G. Lu, S.~Luo, Q.~Zhang, and L.~Zhang, ``Straight line detection algorithm of truck point cloud based on statistical line chart,'' in \emph{Chinese Control and Decision Conference}, 2022, pp. 1206--1211.

\bibitem{ref8}
M.~Poreba and F.~Goulette, ``A robust linear feature-based procedure for automated registration of point clouds,'' \emph{Sensors}, vol.~15, no.~1, pp. 1435--1457, 2015.

\bibitem{ref9}
M.~A. Fischler and R.~C. Bolles, ``Random sample consensus: a paradigm for model fitting with applications to image analysis and automated cartography,'' \emph{Communications of the ACM}, vol.~24, no.~6, pp. 381--395, 1981.

\bibitem{ref10}
M.~Poreba and F.~Goulette, ``Ransac algorithm and elements of graph theory for automatic plane detection in 3d point clouds,'' in \emph{Symposium de PTFiT (Polish Society for Photogrammetry and Remote Sensing)}, vol.~24, 2012, pp. 301--310.

\bibitem{ref11}
P.~Borges, R.~Zlot, M.~Bosse, S.~Nuske, and A.~Tews, ``Vision-based localization using an edge map extracted from 3d laser range data,'' in \emph{IEEE International Conference on Robotics and Automation}, 2010, pp. 4902--4909.

\bibitem{ref12}
R.~Luo, Z.~Zhou, X.~Chu, W.~Ma, and J.~Meng, ``3d deformation monitoring method for temporary structures based on multi-thread lidar point cloud,'' \emph{Measurement}, vol. 200, p. 111545, 2022.

\bibitem{ref13}
P.~Moghadam, M.~Bosse, and R.~Zlot, ``Line-based extrinsic calibration of range and image sensors,'' in \emph{IEEE International Conference on Robotics and Automation}, 2013, pp. 3685--3691.

\bibitem{ref14}
H.~Hoppe, T.~DeRose, T.~Duchamp, J.~McDonald, and W.~Stuetzle, ``Surface reconstruction from unorganized points,'' in \emph{Proceedings of the 19th annual conference on computer graphics and interactive techniques}, 1992, pp. 71--78.

\bibitem{ref15}
X.~Luo, Z.~Tan, and Y.~Ding, ``Accurate line reconstruction for point and line-based stereo visual odometry,'' \emph{IEEE Access}, vol.~7, pp. 185\,108--185\,120, 2019.

\bibitem{ref16}
X.~Zhao, S.~Yang, T.~Huang, J.~Chen, T.~Ma, M.~Li, and Y.~Liu, ``Superline3d: Self-supervised line segmentation and description for lidar point cloud,'' in \emph{European Conference on Computer Vision}, 2022, pp. 263--279.

\bibitem{ref17}
D.~Kong, L.~Xu, X.~Li, and S.~Li, ``K-plane-based classification of airborne lidar data for accurate building roof measurement,'' \emph{IEEE Transactions on Instrumentation and Measurement}, vol.~63, no.~5, pp. 1200--1214, 2013.

\bibitem{ref18}
X.~Lin, Y.~Zhou, Y.~Liu, and C.~Zhu, ``Effective and efficient line segment detection for visual measurement guided by level lines,'' \emph{IEEE Transactions on Instrumentation and Measurement}, 2023.

\bibitem{ref19}
X.~Du, Y.~Lu, and Q.~Chen, ``A fast multiplane segmentation algorithm for sparse 3-d lidar point clouds by line segment grouping,'' \emph{IEEE Transactions on Instrumentation and Measurement}, vol.~72, pp. 1--15, 2023.

\bibitem{ref20}
X.~Lu, Y.~Liu, and K.~Li, ``Fast 3d line segment detection from unorganized point cloud,'' \emph{arXiv preprint arXiv:1901.02532}, 2019.

\bibitem{ref22}
J.~J. Mor{\'e}, ``The levenberg-marquardt algorithm: implementation and theory,'' in \emph{Numerical analysis: proceedings of the biennial Conference held at Dundee, June 28--July 1, 1977}, 2006, pp. 105--116.

\bibitem{ref23}
Z.~Ye, Z.~Wang, X.~Chen, T.~Zhou, C.~Yu, J.~Guo, and J.~Li, ``Zpvehicles: a dataset of large vehicle 3d point cloud data,'' in \emph{IEEE International Workshop on Metrology for Automotive}, 2023, pp. 234--239.

\bibitem{ref21}
J.~Kang and N.~L. Doh, ``Full-dof calibration of a rotating 2-d lidar with a simple plane measurement,'' \emph{IEEE Transactions on Robotics}, vol.~32, no.~5, pp. 1245--1263, 2016.

\bibitem{pointnet}
C.~R. Qi, H.~Su, K.~Mo, and L.~J. Guibas, ``Pointnet: Deep learning on point sets for 3d classification and segmentation,'' in \emph{Proceedings of the IEEE conference on computer vision and pattern recognition}, 2017, pp. 652--660.

\bibitem{pointmlp}
X.~Ma, C.~Qin, H.~You, H.~Ran, and Y.~Fu, ``Rethinking network design and local geometry in point cloud: A simple residual mlp framework,'' \emph{arXiv preprint arXiv:2202.07123}, 2022.

\bibitem{2024survey}
M.~Drobnitzky, J.~Friederich, B.~Egger, and P.~Zschech, ``Survey and systematization of 3d object detection models and methods,'' \emph{The Visual Computer}, vol.~40, no.~3, pp. 1867--1913, 2024.

\bibitem{wang2019research}
Z.~Wang, B.~Yu, J.~Chen, C.~Liu, K.~Zhan, X.~Sui, Y.~Xue, and J.~Li, ``Research on lidar point cloud segmentation and collision detection algorithm,'' in \emph{2019 6th International Conference on Information Science and Control Engineering (ICISCE)}, 2019, pp. 475--479.

\bibitem{subobb}
D.-C. Hoang, L.-C. Chen, and T.-H. Nguyen, ``Sub-obb based object recognition and localization algorithm using range images,'' \emph{Measurement Science and Technology}, vol.~28, no.~2, p. 025401, 2016.

\bibitem{chen2019novel}
L.-C. Chen and T.-H. Nguyen, ``A novel surface descriptor for automated 3-d object recognition and localization,'' \emph{Sensors}, vol.~19, no.~4, p. 764, 2019.

\bibitem{pcl}
R.~B. Rusu and S.~Cousins, ``3d is here: Point cloud library (pcl),'' in \emph{2011 IEEE international conference on robotics and automation}.\hskip 1em plus 0.5em minus 0.4em\relax IEEE, 2011, pp. 1--4.

\bibitem{ptv3}
X.~Wu, L.~Jiang, P.-S. Wang, Z.~Liu, X.~Liu, Y.~Qiao, W.~Ouyang, T.~He, and H.~Zhao, ``Point transformer v3: Simpler faster stronger,'' in \emph{Proceedings of the IEEE/CVF Conference on Computer Vision and Pattern Recognition}, 2024, pp. 4840--4851.

\bibitem{adapointr}
X.~Yu, Y.~Rao, Z.~Wang, J.~Lu, and J.~Zhou, ``Adapointr: Diverse point cloud completion with adaptive geometry-aware transformers. arxiv 2023,'' \emph{arXiv preprint cs.CV/2301.04545}, 2023.

\bibitem{pointcformer}
Y.~Zhong, W.~Quan, D.-m. Yan, J.~Jiang, and Y.~Wei, ``Pointcformer: a relation-based progressive feature extraction network for point cloud completion,'' \emph{arXiv preprint arXiv:2412.08421}, 2024.

\bibitem{cui}
S.~Cui, N.~Wu, and P.~Maghoul, ``Fatigue crack localisation based on empirical mode decomposition and pre-selected entropy,'' \emph{Nondestructive Testing and Evaluation}, pp. 1--28, 2023.

\end{thebibliography}

\end{document}